\documentclass[11pt]{article}

\usepackage[margin=1in]{geometry}
\usepackage{authblk}
\usepackage[hyphens]{url}  
\usepackage{graphicx} 
\usepackage{natbib}  
\usepackage{caption} 
\usepackage{algorithm}
\usepackage{algorithmic}
\usepackage{tikz}
\usetikzlibrary{arrows.meta}
\usepackage{bold-extra}

\usepackage{amsmath}
\usepackage{newfloat}
\usepackage{listings}
\DeclareCaptionStyle{ruled}{labelfont=normalfont,labelsep=colon,strut=off} 
\floatstyle{ruled}
\newfloat{listing}{tb}{lst}{}
\floatname{listing}{Listing}
\usepackage{amsfonts}
\usepackage{amsopn}
\definecolor{keywordcolor}{rgb}{0.7, 0.1, 0.1}
\definecolor{tacticcolor}{rgb}{0.0, 0.1, 0.6}
\definecolor{commentcolor}{rgb}{0.4, 0.4, 0.4}
\definecolor{symbolcolor}{rgb}{0.0, 0.1, 0.6}
\definecolor{sortcolor}{rgb}{0.1, 0.5, 0.1}
\definecolor{stringcolor}{rgb}{0.5, 0.3, 0.0}
\newcommand{\leancolor}{\color}

\lstnewenvironment{leancode}[1][]{
    \lstset{language=lean}
    \lstset{escapeinside={(*}{*)}, mathescape=false}
    #1
  }{}
\definecolor{leanrepaircolor}{rgb}{0.0, 0.75, 0.25}
\def\beginleanrepairAux{\begingroup\boldmath\bfseries\color{leanrepaircolor}\renewcommand{\leancolor}[2][]{\color{leanrepaircolor}}}
\def\beginleanrepair{\aftergroup\beginleanrepairAux}
\def\endleanrepair{\aftergroup\endgroup}
\newcommand{\lean}{\lstset{language=lean,mathescape=false}\lstinline}
\usepackage{amssymb}
\usepackage[table]{xcolor}
\usepackage{booktabs,array,float,longtable,makecell,multirow,subfig,tcolorbox}

\newenvironment{statement}{
\leavevmode\par
\begin{tcolorbox}[
  width=\linewidth,
  colback=gray!5,
  colframe=gray!50,
  boxrule=0.5pt,
  arc=2pt,
  left=6pt,
  right=6pt,
  top=5pt,
  bottom=5pt,
  before skip=5pt,
  after skip=5pt
]
\small}{\end{tcolorbox}}

\usepackage{booktabs}

\title{\textsc{MechGeo}: Autoformalizing and Proving Euclidean Geometry in Lean~4}

\author[1]{Hao Shen\textsuperscript{*}}
\author[2]{Junyu Guo\textsuperscript{*}}
\author[3]{Tian Cui}
\author[1]{Yuxuan Xiao}
\author[1]{Lihong Zhi\textsuperscript{\textdagger}}

\affil[1]{SKLMS,
Academy of Mathematics and Systems Science,
Chinese Academy of Sciences, 
University of Chinese Academy of Sciences,  Beijing 100190, China}

\affil[2]{Institute of Logic and Cognition,
Department of Philosophy, Sun Yat-sen University,
Guangdong 510275, China}

\affil[3]{School of Mathematics, Shandong University,
Shandong 250100, China}

\affil[ ]{%
\textsuperscript{*}Equal contribution.
\quad
\textsuperscript{\textdagger}Corresponding author.%
}

\affil[ ]{%
\mbox{shenhao24@amss.ac.cn};
\mbox{guojy228@mail2.sysu.edu.cn};
\mbox{ct@mail.sdu.edu.cn}%
}

\affil[ ]{%
\mbox{yuxuanxiao@mail.sdu.edu.cn};
\mbox{lzhi@mmrc.iss.ac.cn}%
}
\date{}

\begin{document}

\maketitle

\begin{abstract}

We present \textsc{MechGeo}, a Mathlib native agentic framework that jointly
addresses faithful autoformalization and certified proof construction for
 Euclidean geometry. In this framework, \textsc{GeoFormalizer} represents informal problems in \textsc{GeoIR},
deterministically translates them into Lean~4, and iteratively repairs candidate statements using structural diagnostics and semantic evaluation.
\textsc{GeoProver} constructs geometric proof plans, derives intermediate lemmas, and selectively algebraizes suitable subgoals through a library verified in Lean. Singular or SymPy may generate algebraic certificates, but all resulting proofs and counterexamples are checked by Lean’s kernel. Experiments across seven LLM backbones show substantial improvements in
autoformalization, particularly for models with weaker direct translation performance. On 43 historical IMO geometry problems, \textsc{GeoFormalizer} generates
formal statements that \textsc{GeoProver} proves in 29 cases; for the
remaining 14, it constructs counterexamples verified in Lean and proves all
repaired statements after expert correction. Together with IMO 2026 Problem~2, this yields, to the best of our knowledge, the largest reported collection of automated, kernel-checked Lean proofs for IMO geometry problems. On the 14 geometry statements in \textsc{LEAP}'s Lean-IMO-Bench, \textsc{MechGeo} proves 12 for the first time, formally refutes the remaining two,
and proves both repaired statements. These results establish counterexample guided diagnosis, geometric reasoning, and certified symbolic computation as a practical foundation for trustworthy formal geometry.
\end{abstract}

\section{Introduction}
\tikzset{
  file icon/.pic={
    \begin{scope}[line width=0.85pt]
      \draw[rounded corners=2pt, fill=white] (0.92,3.61) rectangle (2.08,4.62);
      \draw[rounded corners=2pt, fill=white] (0.78,3.73) rectangle (1.94,4.74);
      \draw[rounded corners=2pt, fill=white] (0.64,3.85) rectangle (1.80,4.86);
      \draw (0.90,4.59) -- (1.52,4.59);
      \draw (0.90,4.38) -- (1.52,4.38);
      \draw (1.00,4.05) -- (1.20,4.30) -- (1.40,4.05) -- (1.60,4.30);
      \draw (0.94,4.14) circle (0.105);
    \end{scope}
  },
  module/.style={rectangle, draw=black, rounded corners=5pt, line width=.85pt, fill=white},
  gate/.style={rectangle, draw=black, rounded corners=4pt, line width=.85pt, fill=white},
  part/.style={rectangle, densely dashed, draw=black!50, fill=black!15, minimum height=.5cm, line width=.85pt},
  flow/.style={-{Stealth[scale=1]}, line width=1pt, draw=black},
  wire/.style={-{Stealth[scale=1]}, densely dashed, line width=.7pt, draw=black!50}
}

\definecolor{stageA}{RGB}{250,250,216}   
\definecolor{stageB}{RGB}{239,247,222}   
\definecolor{docBlue}{RGB}{0,150,210}    
\definecolor{docPink}{RGB}{150,35,100}   

\begin{figure*}[t]
    \centering
    \scriptsize
    \resizebox{\columnwidth}{!}{
    \begin{tikzpicture}
    \node[rectangle,minimum width=17.4cm,minimum height=3.5cm,fill=stageA] at (8.8,3.6) {};
    \node[rectangle,minimum width=17.4cm,minimum height=3.3cm,fill=stageB] at (8.8,0.2) {};

    \pic[docBlue, scale=0.8] at (0, 0) {file icon};
    \node at (1.1,2.7) {\textit{Statement}};
    \pic[docBlue, scale=0.8] at (7.0, 0) {file icon};
    \node at (8.1,2.7) {\textit{GeoIR}};
    \pic[docPink, scale=0.8] at (9.8, 0) {file icon};
    \node at (10.9,2.7) {\textit{Formal}};
    \pic[docPink, scale=0.8] at (15.2, 0) {file icon};
    \node at (16.3,2.7) {\textit{Theorem}};

    \def\constrx{3.4}
    \def\constry{3.3}
    \node (mod1) at (\constrx,\constry) [module, minimum width=2cm, minimum height=1.4cm] {};
    \node (llm1) at (\constrx,\constry-0.15) [part, minimum width=1.7cm] {LLM};
    \node at (\constrx,\constry+0.4) {\textbf{Constructor}};

    \node (lint) at (5.9,\constry) [gate, minimum width=1.2cm, minimum height=.7cm] {\textbf{Linter}};
    \draw[flow] (mod1.east)--(lint.west);
    \draw[flow] (lint.east)--node[above]{\includegraphics[width=.35cm]{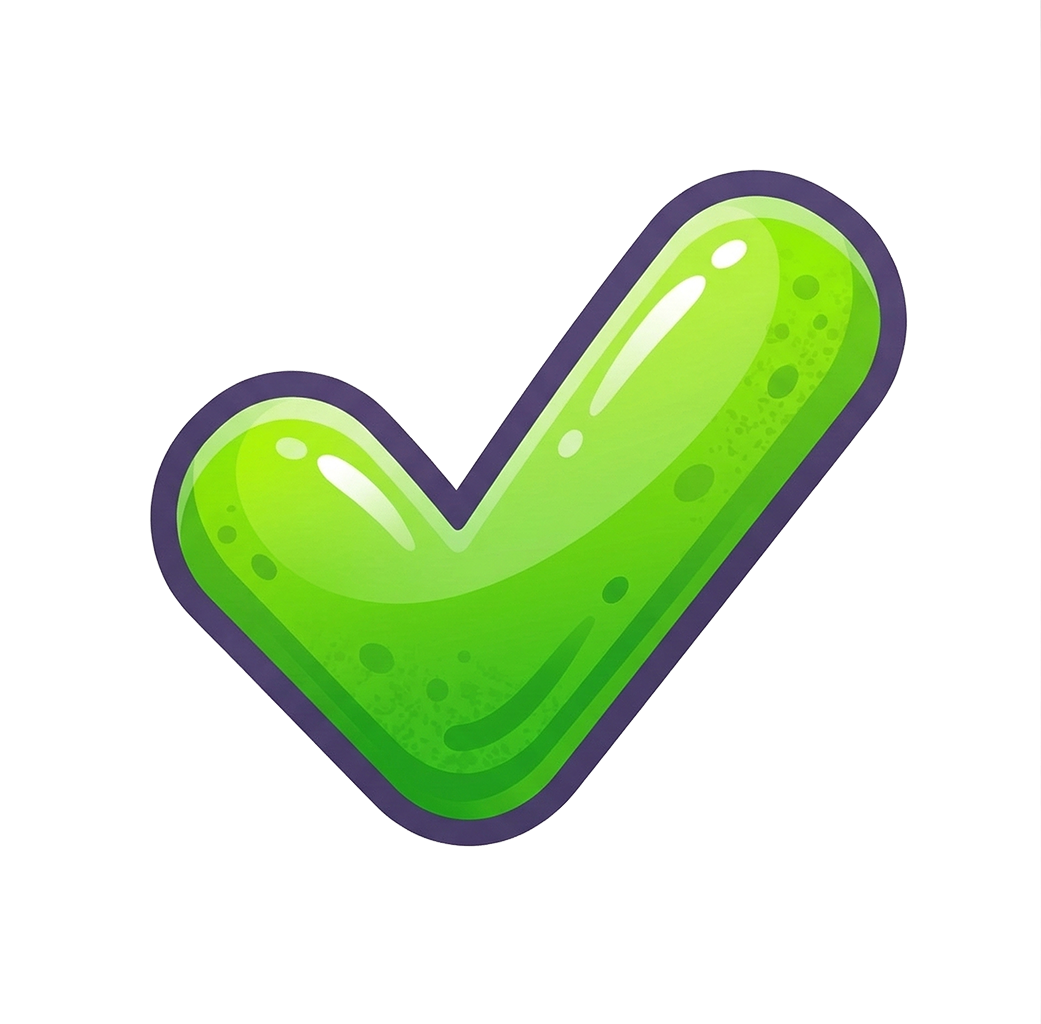}}++(0.9,0);
    \draw[flow] (lint.south)--++(0,-0.95) -| node[pos=0.25, above]{\includegraphics[width=.35cm]{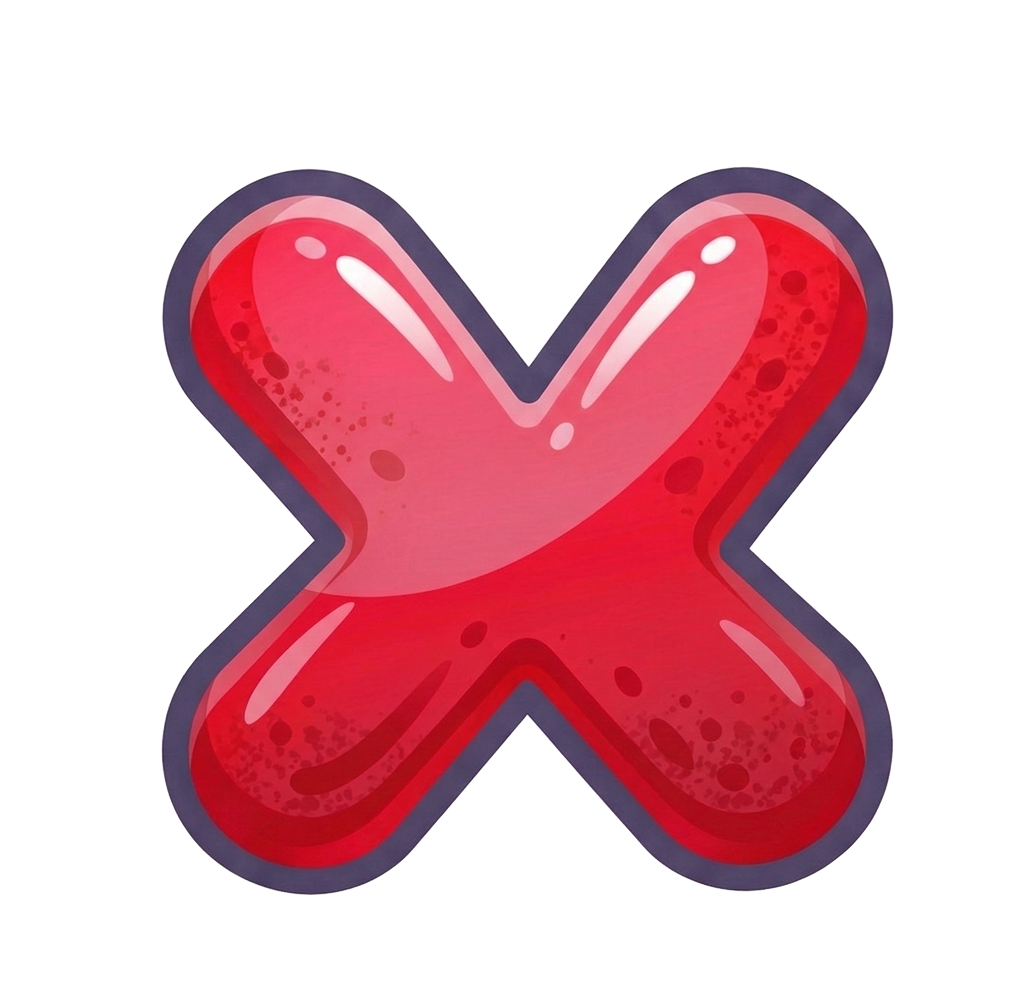}} (mod1.south);

    \def\evalx{13.4}
    \node (mod2) at (\evalx,\constry) [module, minimum width=2cm, minimum height=2cm] {};
    \node (llm2) at (\evalx,\constry+0.15) [part, minimum width=1.7cm] {LLM};
    \node (fscore) at (\evalx,\constry-0.55) [part, minimum width=1.7cm] {F-score};
    \node at (\evalx,\constry+0.7) {\textbf{Evaluator}};

    \draw[flow] ([xshift=-.6cm]mod1.west)--(mod1.west);
    \draw[flow] ([xshift=-.85cm]mod2.west)--(mod2.west);
    \draw[flow] ([xshift=-3.6cm]mod2.west)-- node[above]{\includegraphics[trim=0cm 0.6cm 0cm 0cm, clip=true, width=.4cm]{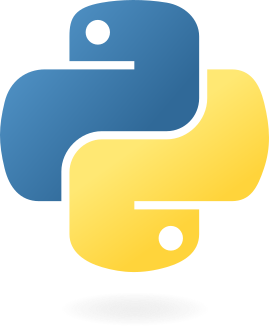}} ([xshift=-2.25cm]mod2.west);
    \draw[flow] (mod2.east)--node[above]{\includegraphics[width=.5cm]{checkmark.png}}([xshift=1.2cm]mod2.east);
    \draw[flow] (mod2.north)--++(0,0.4) -| node[above,xshift=5.0cm]{\includegraphics[width=.5cm]{cross.png}}(mod1.north);

    \pic[docPink, scale=0.8] at (0, -3.2) {file icon};
    \node at (1.1,-0.5) {\textit{Theorem}};
    \pic[docPink, scale=0.8] at (4.8, -3.2) {file icon};
    \node at (5.9,-0.5) {\textit{Subgoals}};
    \pic[docPink, scale=0.8] at (15.2, -3.2) {file icon};
    \node at (16.3,-0.5) {\textit{Proof}};

    \def\planx{3.4}
    \def\plany{0.2}
    \node (mod3) at (\planx,\plany) [module, minimum width=2cm, minimum height=1.4cm] {};
    \node (llm3) at (\planx,\plany-0.15) [part, minimum width=1.7cm] {LLM};
    \node at (\planx,\plany+0.4) {\textbf{Planner}};
    \draw[flow] ([xshift=-.6cm]mod3.west)--(mod3.west);
    \draw[flow] (mod3.east)--([xshift=.8cm]mod3.east);

    \def\lanex{10.45}
    \node (reas) at (\lanex,1.05) [module, minimum width=4.4cm, minimum height=1.3cm] {};
    \node at (\lanex,1.4) {\textbf{Reasoning}};
    \node (mathlib) at (\lanex,0.8) [part, minimum width=3.2cm] {Mathlib lemmas};

    \node (comp) at (\lanex,-0.65) [module, minimum width=4.4cm, minimum height=1.3cm] {};
    \node at (\lanex,-0.3) {\textbf{Computation}};
    \node (agent) at (\lanex-1.1,-0.9) [part, minimum width=1.1cm] {Agent};
    \node (cas) at (\lanex+0.6,-0.9) [part, minimum width=1.1cm] {CAS};
    \node (corr) at (\lanex+1.55,-0.9) [circle, densely dashed, draw=black!50, fill=black!15, line width=.85pt, inner sep=0pt] {\includegraphics[width=.3cm]{checkmark.png}};
    \draw[wire] ([yshift=2.5pt]agent.east)--([yshift=2.5pt]cas.west);
    \draw[wire] ([yshift=-2.5pt]cas.west)--([yshift=-2.5pt]agent.east);
    \draw[-,densely dashed, line width=.7pt, draw=black!50] (cas.east)--(corr.west);

    \draw[wire] ([xshift=-2.5pt]reas.south)--([xshift=-2.5pt]comp.north);
    \draw[wire] ([xshift=2.5pt]comp.north)--([xshift=2.5pt]reas.south);

    \draw[-, line width=1pt, draw=black] (6.55,\plany)--(7.2,\plany);
    \draw[flow] (7.2,\plany)|-node[pos=0.78,above,xshift=-4pt]{\textit{synthetic}}(reas.west);
    \draw[flow] (7.2,\plany)|-node[pos=0.78,below,xshift=-4pt]{\textit{algebraic}}(comp.west);

    \node (verif) at (14.15,\plany) [gate, minimum width=1.35cm, minimum height=.9cm] {\textbf{Verifier}};
    \draw[flow] (reas.east)--++(0.35,0)|-([yshift=0.18cm]verif.west);
    \draw[flow] (comp.east)--++(0.35,0)|-([yshift=-0.18cm]verif.west);
    \draw[flow] (verif.east)--node[above]{\includegraphics[width=.5cm]{checkmark.png}}([xshift=.75cm]verif.east);
\end{tikzpicture}
    }
    \caption{Overview of \textsc{MechGeo}. \textsc{GeoFormalizer} (top, yellow) translates an
informal statement through \textsc{GeoIR} into a linted, compiled, and
semantically evaluated Lean theorem. \textsc{GeoProver} (bottom, green) plans a proof,
decomposes it into subgoals, discharges synthetic subgoals with Mathlib
lemmas and algebraic ones through Lean-checked CAS certificates, and an
independent verifier audits the final proof or counterexample 
artifact. Blue and red
icons denote informal and formal artifacts, respectively.}
    \label{fig:project-overview}
\end{figure*}

Large language models now solve difficult competition problems and are beginning to contribute to mathematical research, while formal reasoning systems increasingly construct machine checked proofs~\citep{kung2026leap,moakhar2026beyond}. Lean~4~\citep{moura2021lean} provides a rigorous foundation by representing mathematical claims precisely and checking proof terms with a small trusted kernel, supported by Mathlib~\citep{mathlib}, a large library of formalized mathematics. Recent systems span specialized provers and agentic frameworks: \textsc{AlphaProof}, \textsc{Seed-Prover}, and \textsc{Aristotle} have demonstrated Olympiad level formal reasoning, while \textsc{DeepSeek-Prover-V2}, \textsc{Goedel-Prover-V2}, and \textsc{Hilbert} have advanced Lean theorem proving~\citep{hubert2025olympiad,chen2025seed,achim2025aristotle,ren2025deepseek,lin2025goedel,varambally2026hilbert}. \textsc{AxiomProver}\footnote{https://github.com/AxiomMath/Putnam2025}, \textsc{Numina-Lean-Agent}, and \textsc{LEAP} have each reported verified Lean solutions to all twelve Putnam 2025 problems~\citep{xin2026axle, liu2026numina,kung2026leap}. Most evaluations, however, begin with already formalized statements, leaving open a prior challenge: faithfully translating the intended configuration into a formal statement. 

Faithful formalization is  challenging in Euclidean geometry.
Informal problems derive their meaning jointly from text, diagrams, and
geometric conventions, often leaving necessary order and nondegeneracy
conditions implicit. Human readers naturally interpret the statement through
the intended diagram and may therefore overlook unintended configurations
admitted by the formal assumptions. Because many Mathlib definitions remain
meaningful on degenerate inputs, such omissions need not cause a compilation
error. For example, ``$D$ lies on $BC$'' may mean that $D$ lies on the
supporting line, on the closed segment, or strictly between $B$ and $C$, with
$B \ne C$ often left implicit. These interpretations yield different Lean
statements, all of which may elaborate, and some may even admit valid proofs.
Thus, neither compiler feedback nor human inspection alone reliably determines
whether a formal statement captures the intended configuration
~\citep{murphy2024autoformalizing,tang2026euclean}.

Proving a formalized theorem poses a second challenge: exposing enough geometric structure for effective proof search. Specialized systems such as \textsc{AlphaGeometry} and \textsc{TongGeometry} achieve strong performance through specialized languages and symbolic deduction, but their success depends on proposing the required auxiliary constructions and expressing the relevant relations~\citep{trinh2024solving,zhang2026proposing}. \textsc{AlphaGeometry2} expands this language to express \(88\%\) of IMO geometry problems from 2000 to 2024; the remaining cases involve inequalities, nonlinear relations, three dimensional configurations, or constructions with arbitrarily many points~\citep{chervonyi2025gold}. Mathlib native formalization is complementary, embedding geometry in a general mathematical library and producing reusable proofs checked by the Lean kernel.

Algebraic reasoning is a central tool for geometry theorem proving. Coordinate algebraization translates geometric hypotheses and goals into polynomial equations and inequalities, which can then be handled systematically by Wu's method, Gr\"obner basis computation, and related symbolic procedures~\citep{wen1986basic,RISC400}.  Although algebraic methods reduce dependence on auxiliary constructions,
algebraizing an entire configuration may produce large polynomial systems
and substantial expression growth, with efficiency sensitive to coordinate
 choice, formulation, and variable ordering.
Geometric reasoning and algebraic computation are therefore complementary: the former preserves geometric structure and decomposes the theorem into manageable subgoals, while the latter systematically discharges the resulting polynomial consequences.

We present \textsc{MechGeo}, a Mathlib native agentic framework for autoformalization and certified Euclidean geometry proving. Its central
principle is that geometric reasoning and algebraic computation should work
together: geometric reasoning exposes the structure of a configuration and
decomposes difficult proofs into manageable subgoals, while our formally
verified algebraization library translates selected obligations into polynomial
form for systematic symbolic computation. Our contributions are threefold:

\begin{itemize}
    \item \textbf{Mathlib native autoformalization.}
 \textsc{GeoFormalizer} introduces \textsc{GeoIR}, a deterministic Lean
 translation, and repair guided by compiler and semantic feedback. 
    Across seven LLM backbones, it substantially improves the success rates of elaboration,
especially for models with weaker direct-translation performance. 
    

    \item \textbf{Certified proving through geometric and algebraic reasoning.}
     \textsc{GeoProver} combines proof planning, intermediate lemmas, and
    selective algebraization. External CAS may generate certificates, but Lean
    verifies every certificate and final proof.

 \item \textbf{Verified IMO geometry results.}
We construct the faithful formalizations and kernel-checked proofs for 44 IMO
geometry problems from 2000 to 2026. 
\textsc{MechGeo} also proves 12 statements in \textsc{LEAP}'s Lean-IMO-Bench~\citep{kung2026leap} for the first time, formally refutes the remaining two, and proves both after repair. The results are
publicly available in the \textsc{MechGeoBench}
repository.\footnote{\url{https://github.com/MechMath/MechGeoBench}}
    
\end{itemize}
\section{Related Work}

\paragraph{Symbolic and Neuro-Symbolic Geometry Reasoning.}
Classical approaches include Wu's method, Gr\"obner basis computation, area
and full angle methods, and deductive databases
~\citep{wen1986basic,RISC400,chou1994machine,chou2000deductive}.
Recent systems such as \textsc{AlphaGeometry}, \textsc{AlphaGeometry2}, and
\textsc{TongGeometry} combine symbolic deduction with learned guidance for
auxiliary construction and search
~\citep{trinh2024solving,chervonyi2025gold,sinha2024wu,
zhang2026proposing}. Unlike these systems, which use specialized geometric
languages, \textsc{MechGeo} works with Mathlib statements and produces proofs
checked by the Lean kernel.

\paragraph{Euclidean Geometry in Proof Assistants.}
Earlier work formalized the area method, a simple version of Wu's method,
the connection between synthetic and Cartesian geometry in \textsc{GeoCoq},
and Euclid's Book~I in Coq and HOL Light
~\citep{narboux2004decision,genevaux2011formalization,
boutry2019formalization,beeson2019proofchecking}. In Lean,
\textsc{LeanEuclid} translates informal problems into System~E and uses SMT
reasoning to recover implicit diagrammatic facts; 
\textsc{LeanGeo} provides a
library, automation, and a benchmark for competition geometry, and 
\textsc{Euclean} targets Mathlib-native geometry statement formalization
through constraint explication, configuration anchoring, formalization
mapping, and iterative repair
~\citep{murphy2024autoformalizing,song2025leangeo,tang2026euclean}.
\textsc{MechGeo} unifies Mathlib native autoformalization, semantic repair,
and certified proof construction through a Lean verified algebraization layer.

\section{Methodology}

\begin{figure*}[t]
\centering
\includegraphics[width=\textwidth]{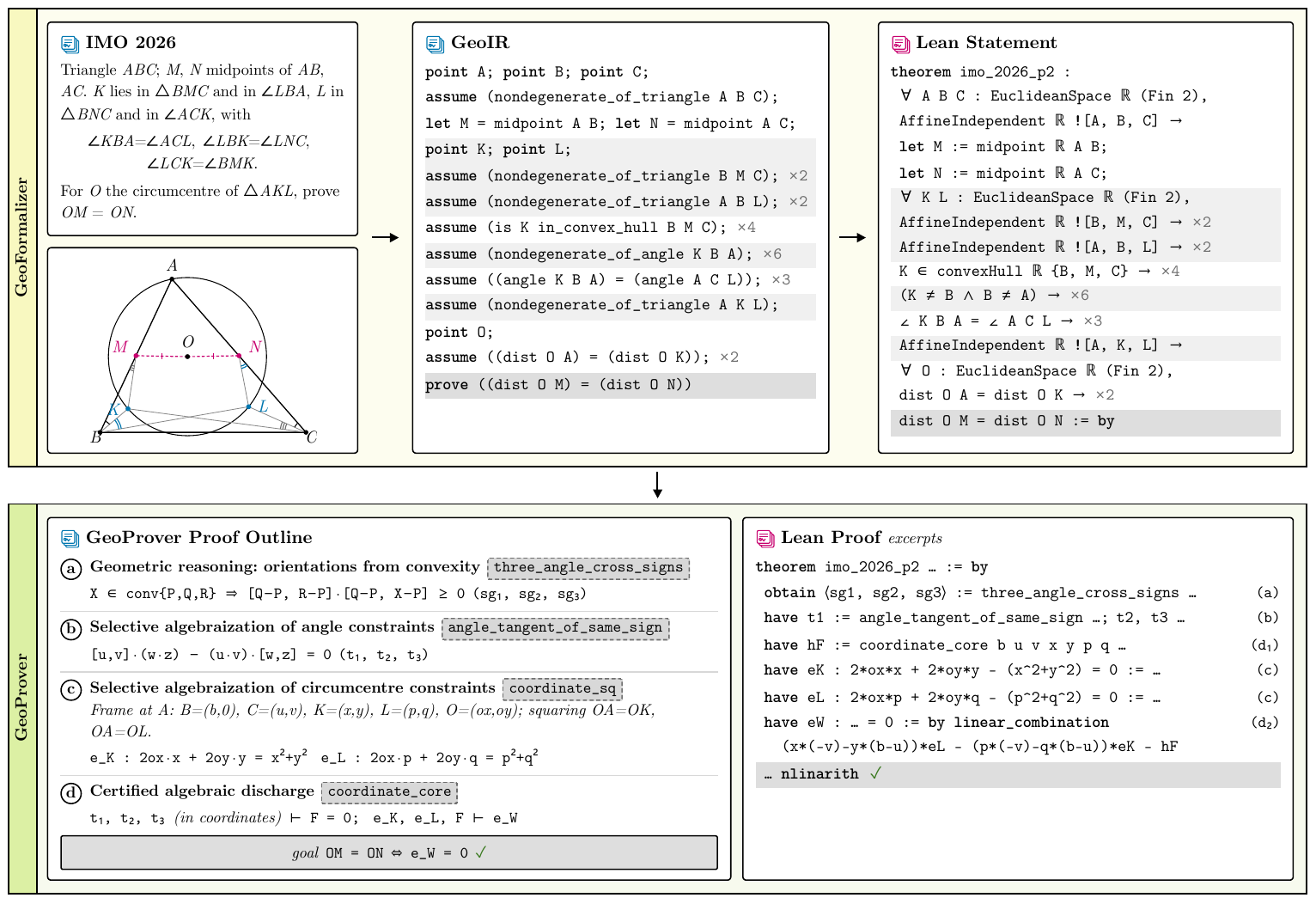}
\caption{Case study on IMO 2026 P2: \textsc{GeoIR} to Lean~4 statement,
then \textsc{GeoProver}'s plan (a)--(d) and the Lean proof lines it maps to.
Repeated hypothesis families are folded as $\times n$.}
\label{fig:imo2026-case-study}
\end{figure*}

In this section, we present \textsc{MechGeo}, a framework that translates
informal Euclidean geometry problems into Mathlib native Lean~4 statements and
constructs complete proofs without \texttt{sorry}. As shown in Figure~\ref{fig:project-overview},
\textsc{GeoFormalizer} generates a checked \textsc{GeoIR} representation,
translates it deterministically into Lean, and repairs it using static, compiler,
and semantic feedback. \textsc{GeoProver} first constructs a geometric proof plan that identifies
auxiliary constructions and intermediate facts. Guided by this plan, it keeps
structural arguments at the Mathlib level and selectively translates suitable
hypotheses and goals into polynomial constraints. This selective algebraization
preserves geometric structure, keeps the resulting polynomial systems
tractable, and allows symbolic methods to discharge exact algebraic
consequences. All generated certificates and the final proof are checked in Lean. Figure~\ref{fig:imo2026-case-study} illustrates this workflow on IMO 2026 Problem 2.

\subsection{\textsc{GeoFormalizer}: Autoformalization via \textsc{GeoIR}}
\label{subsec:method-formalization}
\textsc{GeoFormalizer} comprises four stages: \textsc{GeoIR} generation, Lean~4 statement translation, formal and semantic evaluation, and iterative repair.

\subsubsection{\textsc{GeoIR} Generation}

Directly generating Mathlib native geometry statements requires substantial knowledge of Lean syntax, its type system, and Mathlib interfaces, beyond what is needed to express standard Euclidean geometry problems.
\textsc{GeoIR} reduces this burden by providing a simple, compact, and typed language in which models can state geometric content with a limited set of notations without directly manipulating statements in Lean.

\subsubsection{Formal Statement Translation} 

There are four categories of \textsc{GeoIR} constructs: declarations, constructions, geometric relations, and values. 
Each construct has a fixed translation in Lean. Most of these constructs are translated into corresponding Mathlib expressions, and a small number  are translated into predicates from our library. Translation is performed by a rule-based translator without an LLM. The translator parses a geometry proposition written in \textsc{GeoIR} into a typed abstract syntax tree, recursively traverses the tree, and reconstructs the corresponding Lean statement using a fixed mapping table. Lean's parser, elaborator, and kernel then check well-formedness and well-typedness of the generated statement. The syntax and symbols used in the statement are confined to the translator mappings. When an interface in Mathlib changes, the affected mappings can be updated, and existing \textsc{GeoIR} statements can be re-translated into Lean statements that fit the new interface.

\subsubsection{Evaluation}
A challenge in automated geometry formalization is detecting and
repairing semantic mismatches between a natural language problem and its formal
statement. A statement may type check in Lean yet fail to capture the intended
theorem because its assumptions are inconsistent, its conclusion is trivial, or
it expresses a different theorem. We therefore combine semantic and structural
signals into a single score:
$
F = 0.7 S_{\mathrm{judge}} + 0.3 S_{\mathrm{struct}}
$.
The judge score \(S_{\mathrm{judge}}\) evaluates the correspondence between the
informal problem and the generated Lean statement. It assigns one of three labels: semantically consistent, partially consistent, or inconsistent, which are mapped to scores of \(1\), \(0.5\), and \(0\), respectively. The structural
score $S_{\mathrm{struct}}$ is the point coverage ratio, defined as the
fraction of points declared in the \textsc{GeoIR} specification that are mentioned in the informal problem. 
This scoring design motivates our choice of $\tau=0.6$, so $S_{\mathrm{struct}}$ is decisive only in the ambiguous case $S_{\mathrm{judge}}=0.5$.

\subsubsection{Iterative Repair}
The generated \textsc{GeoIR} specification is first parsed and statically checked for undeclared objects, arity errors, type mismatches, and malformed structures.
Diagnostics from parsing and static checking are returned to the constructor for repair. 
Candidates that  pass these checks receive a score $F$ and are accepted if
$F \geq 0.6$. Otherwise, the evaluator provides semantic and structural
feedback, and the constructor revises the \textsc{GeoIR} specification.

\subsection{\textsc{GeoProver}: Integrating Geometric and Algebraic Reasoning}
\label{sec:method-reasoning}

\textsc{GeoProver} combines geometric and algebraic reasoning in a four stage agentic workflow: (1) \emph{Proof Planning}, (2) \emph{Selective Algebraization}, (3) \emph{Certified Algebraic Discharge}, and (4) \emph{Independent Verification}.

\subsubsection{Proof Planning}
\label{sec:geoprover-planning}
For each theorem, \textsc{GeoProver} gives the proof agent only the formal Lean statement. The agent reconstructs the geometry and organizes key lemmas, auxiliary constructions, angle chases, and algebraic identities into an ordered informal proof plan. This decomposition precedes algebraization, so each CAS call typically handles a smaller, lower degree polynomial subsystem, reducing the burden on symbolic elimination.

\subsubsection{Selective Algebraization}
\label{sec:geoprover-dualmode}

Algebraization is implemented by a collection of verified lemmas and the
simplification tactic \texttt{to\_poly}, which applies some of these lemmas to the
goal or selected hypotheses.
It rewrites geometric notions, including predicates (e.g., collinearity and concyclicity), geometric objects (e.g., points and circles), and real number quantities (e.g., distances and arithmetic expressions), into polynomial expressions, equations, and inequalities in point coordinates. Logically, it proceeds in two steps. First, derived geometric predicates and objects are expressed with basic notations, including vectors, norms, and inner products, as handled by the   tactic \texttt{to\_basic}. Then, these expressions are expanded coordinatewise into polynomials, as done by tactic \texttt{basic\_to\_poly}. The agent may apply selected lemmas or tactics at specialized locations. This selective control allows algebraization on specialized hypotheses and goals, such as subgoals proved with algebraic computation while keeping the remaining part in geometric form, on which existing lemmas in Mathlib for geometry can be applied. Rewrite rules applied by \texttt{to\_poly} are lemmas of equivalence between the geometric expressions and their algebraic expressions, so they never silently strengthen or weaken the statement. Typical proofs alternate between the two levels: synthetic lemmas establish geometric facts that are subsequently used for algebraization.

\subsubsection{Certified Algebraic Discharge}
\label{sec:geoprover-geocas}
For polynomial equality goals with hypotheses $p_1=0,\ldots,p_n=0$ and target $q=0$, when Lean's internal tactics are insufficient, \textsc{GeoProver} calls Singular~\citep{DGPS19} or SymPy~\citep{meurer2017sympy} to compute cofactors $c_1,\ldots,c_n$ that satisfy $q=\sum_{i=1}^{n}c_i p_i.$
The agent translates this identity into a Lean proof, which is checked by
the kernel. Inequalities and nonzero constraints are handled using Lean
tactics, explicit sign lemmas, and case analysis. A CAS may assist with
factorization or algebraic decomposition, but all sign arguments are verified in Lean.

\paragraph{Independent Verification.}
For each input statement, \textsc{GeoProver} attempts either to prove it or to
refute it with a  counterexample. In the latter case, explicit coordinates are usually used to construct a counterexample and prove that the conclusion can fail when hypotheses hold. An independent verifier checks whether the statement has
been correctly proved or refuted: a proof must match the input statement
exactly, whereas a counterexample must formally establish its negation.  
In either case, the complete file must compile without \texttt{sorry}
and use only the permitted Lean axioms.

\section{Experiments}
\label{sec:experiments}

We evaluate \textsc{GeoFormalizer} and \textsc{GeoProver} both separately and
within the complete \textsc{MechGeo} pipeline. End-to-end accuracy alone
conflates three distinct failure modes: the generated statement may fail to
elaborate, may elaborate but misrepresent the source problem, or may not yield
a kernel-verified proof. We therefore report elaboration, semantic consistency,
and proof success separately, organized around three research questions:

\textbf{RQ1. Autoformalization.}
How accurately does \textsc{GeoFormalizer} generate Lean statements that elaborate, and how much do iterative repairs improve performance?

\textbf{RQ2. Theorem proving and refutation.}
Given a fixed Lean statement, how often does \textsc{GeoProver} either construct
a kernel-checked proof or formally refute it with a Lean-verified
counterexample, and how much do selective algebraization and external computer
algebra contribute? 

\textbf{RQ3. Olympiad-level reasoning.}
How broadly can \textsc{MechGeo} prove or refute Olympiad-level geometry
statements? We evaluate \textsc{MechGeo} on 44 human audited IMO formalizations, 
repaired where necessary, as well as all 14 geometry problems in Lean-IMO-Bench~\citep{kung2026leap}.

\subsection{Experimental Settings}
\label{sec:exp-settings}
\subsubsection{Dataset}

Our evaluation set, \textsc{MG200}, contains four subsets totaling 200
informal geometry problems: 43 historical IMO geometry problems, 77 non-IMO
problems from LeanGeo-Bench~\citep{song2025leangeo}, 22 classical geometry theorems, denoted \textsc{CertiGeo}, used to evaluate
algebraic proving~\citep{gregoire2011proofcertificates}, 
and 58 plane geometry problems sampled by
difficulty from \textsc{Euclidean's} OMNI-Geometry dataset
~\citep{tang2026euclean}.
  Together, these
subsets are used to evaluate \textsc{GeoFormalizer}. To isolate proof generation from statement formalization, we evaluate \textsc{GeoProver} on the Lean statements produced by \textsc{GeoFormalizer} for all 200 \textsc{MG200} problems.
We further test \textsc{GeoProver} on 31 plane Euclidean
geometry statements based on \lean{EuclideanSpace ℝ (Fin 2)} in Lean selected from PutnamBench
~\citep{tsoukalas2024putnambench} and on all 14 geometry problems in
Lean-IMO-Bench, introduced by \textsc{LEAP}~\citep{kung2026leap}, to assess transfer beyond the main benchmark and performance on IMO-level geometry.

\subsubsection{Models and Tools}
We evaluate autoformalization with seven strong LLM backbones
from diverse providers: GPT-5.6-Sol~\citep{openai2026gpt56sol},
Claude Opus 4.8~\citep{anthropic2026claudeopus48},
DeepSeek-V4-Pro and DeepSeek-V4-Flash~\citep{deepseekai2026deepseekv4},
Qwen3.7-Max~\citep{qwenteam2026qwen37max},
MiniMax-M3~\citep{minimax2026m3}, and
GLM-5.2~\citep{zai2026glm52}.
The selection includes both proprietary and open-weight model
families, allowing us to assess whether the benefits of
\textsc{GeoFormalizer} generalize beyond any single model or provider.
All experiments use Lean~4.27.0. External symbolic computation uses Singular~4.4.1 and SymPy~1.13.2.

\subsection{RQ1: Automatic Formalization}
\label{sec:exp-autoformalization}
\begin{table*}[!t]
\centering
{\small
\setlength{\tabcolsep}{1mm}
\renewcommand{\arraystretch}{1.08}
\begin{tabular*}{\textwidth}{
  @{\extracolsep{\fill}}
  llccccccc
  @{}
}
\toprule
Dataset
& Method
& \shortstack{GPT-5.6\\Sol}
& \shortstack{Claude\\Opus 4.8}
& \shortstack{DeepSeek\\V4-Pro}
& \shortstack{DeepSeek\\V4-Flash}
& \shortstack{Qwen\\3.7-Max}
& \shortstack{MiniMax\\M3}
& \shortstack{GLM\\5.2} \\
\midrule

\multirow{3}{*}{\textsc{IMO} (43)}
& Direct
& 21 (48.8)
& 23 (53.5)
& 1 (2.3)
& 0 (0.0)
& 30 (69.8)
& 0 (0.0)
& 16 (37.2) \\

& Euclean
& 43 (100)
& 43 (100)
& \textbf{43 (100)}
& \textbf{43 (100)}
& 31 (72.1)
& 20 (46.5)
& 29 (67.4) \\

& Ours
& \textbf{43 (100)}
& \textbf{43 (100)}
& 39 (90.7)
& 42 (97.7)
& \textbf{39 (90.7)}
& \textbf{41 (95.3)}
& \textbf{43 (100)} \\

\addlinespace[2pt]

\multirow{3}{*}{\textsc{CertiGeo} (22)}
& Direct
& 16 (72.7)
& 15 (68.2)
& 4 (18.2)
& 4 (18.2)
& 17 (77.3)
& 3 (13.6)
& 12 (54.5) \\

& Euclean
& 22 (100)
& 22 (100)
& 22 (100)
& 22 (100)
& 16 (72.7)
& 10 (45.5)
& 13 (59.1) \\

& Ours
& \textbf{22 (100)}
& \textbf{22 (100)}
& \textbf{22 (100)}
& \textbf{22 (100)}
& \textbf{22 (100)}
& \textbf{21 (95.5)}
& \textbf{22 (100)} \\

\addlinespace[2pt]

\multirow{3}{*}{
  \shortstack[l]{\textsc{LeanGeo}\\Non-IMO (77)}
}
& Direct
& 33 (42.9)
& 25 (32.5)
& 5 (6.5)
& 7 (9.1)
& 59 (76.6)
& 9 (11.7)
& 42 (54.5) \\

& Euclean
& 77 (100)
& 77 (100)
& 75 (97.4)
& 77 (100)
& 65 (84.4)
& 37 (48.1)
& 66 (85.7) \\

& Ours
& \textbf{77 (100)}
& \textbf{77 (100)}
& \textbf{76 (98.7)}
& \textbf{77 (100)}
& \textbf{72 (93.5)}
& \textbf{74 (96.1)}
& \textbf{77 (100)} \\

\addlinespace[2pt]

\multirow{3}{*}{
  \shortstack[l]{\textsc{OMNI}\\Geometry (58)}
}
& Direct
& 40 (69.0)
& 35 (60.3)
& 6 (10.3)
& 15 (25.9)
& 37 (63.8)
& 5 (8.6)
& 33 (56.9) \\

& Euclean
& 58 (100)
& 58 (100)
& \textbf{57 (98.3)}
& \textbf{54 (93.1)}
& 28 (48.3)
& 24 (41.4)
& 27 (46.6) \\

& Ours
& \textbf{58 (100)}
& \textbf{58 (100)}
& 48 (82.8)
& 53 (91.4)
& \textbf{53 (91.4)}
& \textbf{53 (91.4)}
& \textbf{58 (100)} \\

\bottomrule
\end{tabular*}
}
\caption{
RQ1 results on the four autoformalization datasets. Ours denotes \textsc{GeoFormalizer}. Each entry reports the number (percentage) of successfully elaborated Lean statements.
Direct uses a single translation attempt. Although the evaluation settings differ, Euclean receives a 10-minute timeout and, in our runs, makes 14--116 model calls per problem, whereas
 \textsc{GeoFormalizer} uses at most five calls.
Best results within each dataset--model column are bolded.
}
\label{tab:rq1-main}
\end{table*}

In all RQ1 experiments, the automatic semantic judge used for semantic-guided repair is fixed to \textsc{GPT-5.6-Sol}
across all constructor backbones. Table~\ref{tab:rq1-main} shows that the structured formalization pipeline
substantially improves elaboration rate across model families. Across all
1,400 model–problem pairs, \textsc{GeoFormalizer} produces 1,354 statements
that elaborate successfully (96.7\%), compared with 1,159 (82.8\%) for
\textsc{Euclean} and 513 (36.6\%) for direct translation. The gains over
\textsc{Euclean} are particularly large for backbones with weaker baseline
performance: the overall elaboration rate rises from 70.0\% to 93.0\% for
Qwen3.7-Max, from 45.5\% to 94.5\% for MiniMax-M3, and from 67.5\% to
100\% for GLM-5.2. \textsc{GeoFormalizer} also preserves near-ceiling
performance for the strongest backbones, achieving 100\% with GPT-5.6-Sol and Claude Opus 4.8. The improvements are consistent across datasets. Even on IMO and OMNI-Geometry, where several baselines degrade most sharply, \textsc{GeoFormalizer} maintains strong performance across backbones, with most results exceeding 90\%. 
Although the evaluation settings differ, \textsc{Euclean} uses a substantially
larger model call budget: in our runs, it makes 14--116 model calls per problem,
whereas \textsc{GeoFormalizer} uses at most five. 

\begin{table}[t]
\centering
\footnotesize
\renewcommand{\arraystretch}{1.05}

\begin{tabular*}{\columnwidth}{
  @{\extracolsep{\fill}}lccccc@{}
}
\toprule
Model & \multicolumn{4}{c}{Setting $(r,f)$} & $\Delta$ \\
\cmidrule(lr){2-5}
& $(0,0)$ & $(0,2)$ & $(2,0)$ & $(2,2)$ & \\
\midrule
Claude Opus 4.8
  & 98.5 & 99.0 & 97.5 & \textbf{100.0} & +1.5 \\
GLM-5.2
  & 97.5 & 98.0 & 100.0 & \textbf{100.0} & +2.5 \\
GPT-5.6 
  & 97.5 & 96.0 & 100.0 & \textbf{100.0} & +2.5 \\
DeepSeek V4-Flash
  & 85.0 & 87.0 & 97.0 & \textbf{97.0} & +12.0 \\
DeepSeek V4-Pro
  & 81.0 & 85.0 & \textbf{96.5} & 92.5 & +11.5 \\
Qwen3.7 Max
  & 64.5 & 63.0 & 90.5 & \textbf{93.0} & +28.5 \\
MiniMax M3
  & 57.5 & 93.0 & \textbf{96.5} & 94.5 & +37.0 \\
\midrule
Mean
  & 83.1 & 88.7 & 96.9 & 96.7 & +13.6 \\
\bottomrule
\end{tabular*}
\caption{RQ1 ablation of repair rounds on the 200-problem benchmark. Each cell reports the percentage of problems whose generated Lean statement elaborates successfully under setting $(r,f)$, where $r$ and $f$ denote the numbers of compiler-guided and semantic-guided repair rounds, respectively. $\Delta$ is the absolute improvement of the full setting $(2,2)$ over the baseline $(0,0)$.}
\label{tab:rq1-ablation}
\end{table}

Table~\ref{tab:rq1-ablation} further examines the contributions of compiler
guided and semantic-guided repair in \textsc{GeoFormalizer}. We compare
four settings by independently enabling or disabling the two repair components.
Without repair, the mean elaboration rate is 83.1\%. Two rounds of compiler
guided repair increase it to 96.9\%, whereas semantic-guided repair alone
raises it to 88.7\%. Thus, compiler guided repair contributes most of the gain
in elaboration rate. Semantic guided repair has a smaller effect on elaboration because it
primarily targets semantic mismatches.  

Because Lean acceptance guarantees only syntactic and type correctness, three
experts independently evaluate the semantic consistency of the 200 statements
generated by Claude Opus 4.8. A statement is accepted by majority vote only if
it preserves the intended objects, constructions, hypotheses, admissible
configurations, and conclusion, allowing equivalent reformulations. Initially,
157 statements are judged faithful, giving a rate of 78.5\%. However,
Table~\ref{tab:human-formal-crosscheck} shows that formal counterexample search
refutes 22 of these statements, revealing degeneracy and order errors missed
by human inspection.

\subsection{RQ2: Automated Theorem Proving and Refutation }
\label{sec:exp-theorem-proving}

For RQ2, we evaluate all provers on the 200 statements generated by
\textsc{GeoFormalizer} with Claude Opus 4.8, including statements later judged
semantically unfaithful, along with 31 Euclidean geometry statements in Lean from
PutnamBench. Each prover receives a two hour budget per problem and runs in a
sandbox with read only access to the Lean project and toolchain, a private
writable workspace, and no web access. \textsc{Numina-Lean-Agent} and
\textsc{Hilbert} use the same DeepSeek-V4-Pro backend, input statements, and
budget as \textsc{GeoProver}. For both agents, we retain the default prompts
from their official releases and replace only the LLM backend.
\textsc{Goedel-Prover-V2-8B} uses its released 8B model with pass@32 sampling,
following its original evaluation protocol. As shown in Table~\ref{tab:rq2-main-ablation},
\textsc{GeoProver} proves 67 of the 200 MG200 statements and 6 of the 31
PutnamBench statements in Lean, for a total of 73 out of 231, compared with 44 for
\textsc{Numina-Lean-Agent}, 13 for \textsc{Hilbert}, and none for
\textsc{Goedel-Prover-V2-8B}.

To evaluate both proof construction and formal refutation,
Table~\ref{tab:human-formal-crosscheck} presents  the cross-tabulation between the results on \textsc{MG200} and human semantic annotations.
Replacing the backbone with GPT-5.6-Sol while keeping all other settings unchanged substantially improves performance on both datasets. On MG200, \textsc{GeoProver} proves 145 of the 200 statements and formally refutes 45. On the Putnam subset, it proves 30 of the 31 statements in Lean, with only one timeout.  The results further illustrate that successfully proving a formal statement does not necessarily imply semantic consistency with the original informal problem. Moreover, because Euclidean geometry problems often rely on implicit nondegeneracy conditions that are difficult for human annotators to verify exhaustively, establishing statement faithfulness remains a challenging problem. We investigate this issue in detail in RQ3.

\begin{table}[!t]
\centering
\small
\renewcommand{\arraystretch}{1.08}
\setlength{\tabcolsep}{0pt}

\begin{tabular*}{\columnwidth}{
  @{\extracolsep{\fill}}
  l
  c
  c
  c
  @{}
}
\toprule
Method / Setting
& \textsc{MG200} (200)
& Putnam (31)
& Total (231) \\
\midrule

Goedel
& 0 (0.0\%)
& 0 (0.0\%)
& 0 (0.0\%) \\

Hilbert
& 13 (6.5\%)
& 0 (0.0\%)
& 13 (5.6\%) \\

Numina
& 43 (21.5\%)
& 1 (3.2\%)
& 44 (19.0\%) \\

\midrule

\textsc{GeoProver} (full)
& \textbf{67 (33.5\%)}
& \textbf{6 (19.4\%)}
& \textbf{73 (31.6\%)} \\

\quad w/o CAS
& 29 (14.5\%)
& 0 (0.0\%)
& 29 (12.6\%) \\

\quad w/o CAS + algebra
& 19 (9.5\%)
& 0 (0.0\%)
& 19 (8.2\%) \\

\bottomrule
\end{tabular*}

\caption{
RQ2 main comparison and cumulative ablation results.
Hilbert, Numina-Lean-Agent, \textsc{GeoProver}, and the \textsc{GeoProver} ablations
use DeepSeek-V4-Pro under the same two-hour per-problem budget.
Goedel-Prover-V2-8B is evaluated with its native model using pass@32.
Each entry reports the number and percentage of statements for which
a kernel-checked proof is obtained.
}
\label{tab:rq2-main-ablation}
\end{table}

\begin{table}[t]
\centering
\small
\renewcommand{\arraystretch}{1.08}
\setlength{\tabcolsep}{0pt}
\begin{tabular*}{\columnwidth}{
@{\extracolsep{\fill}}
l
c
c
c
c
@{}
}
\toprule
& Proved
& Refuted
& Timeout
& Total \\
\midrule
Faithful (Human)
& 128
& 22
& 7
& 157 \\
Not Faithful (Human)
& 17
& 23
& 3
& 43 \\
\midrule
Total
& 145
& 45
& 10
& 200 \\
\bottomrule
\end{tabular*}
\caption{%
Cross-tabulation of initial human semantic judgments and \textsc{GeoProver}
outcomes on MG200 using GPT-5.6-Sol. The statements are generated
by \textsc{GeoFormalizer} with Claude Opus 4.8.}
\label{tab:human-formal-crosscheck}
\end{table}

\subsubsection{Contributions of CAS}
To quantify the contribution of symbolic computations, we perform an ablation study by progressively removing the external computer algebra systems and the Lean-side algebraization toolkit. 
All other components, including the DeepSeek-V4-Pro backbone, remain unchanged. As shown in Table~\ref{tab:rq2-main-ablation}, the full \textsc{GeoProver} achieves 73 verified proofs out of 231 problems. 
Removing CAS causes a substantial performance degradation, reducing the number of solved problems to 29. This demonstrates that external symbolic computation plays a central role in discharging algebraically structured geometric goals. Removing both CAS and the algebraization toolkit further decreases performance to only 19 solved problems.  The additional drop indicates that the algebraization interface is also essential: even without external symbolic solvers, exposing geometric relations as polynomial constraints provides useful intermediate representations for Lean-based reasoning. These results confirm that \textsc{GeoProver's} effectiveness does not come solely from LLM-based proof search. Instead, the combination of geometric reasoning, Lean-verified algebraization, and symbolic computation provides complementary capabilities.

\subsubsection{Qualitative analysis of proof decomposition.}
The ablation results quantify the aggregate contribution of the algebraization and CAS components. We further examine how the agent's planning and decomposition reduce the complexity of individual algebraic goals. For IMO 2008 P1, one of the 73 solved problems, decomposition replaces a global system of 16 quadratic equations in 22 scalar coordinate variables with six local systems, each comprising three quadratic equations and one quadratic nonvanishing constraint in 10 variables, followed by two symmetry systems with three and two quadratic equations, respectively, in at most 10 variables.

 \subsection{RQ3: IMO-Level Geometry Formalization and Proving}
\begin{table}[t]
\centering
\small
\renewcommand{\arraystretch}{1.08}
\setlength{\tabcolsep}{3pt}

\begin{tabular*}{\columnwidth}{
  @{\extracolsep{\fill}}
  l
  c
  c
  c
  @{}
}
\toprule

& \shortstack{\textsc{IMO} (43)}
& \shortstack{IMO 2026}
& \shortstack{\textsc{LEAP}} \\
\midrule

Proved
& 29
& 1
& 12 \\

Refuted
& 14
& 0
& 2 \\

Human repaired 
& 14
& 0
& 2 \\

Repaired proved
& 14
& 0
& 2 \\

\midrule

\textbf{Final faithful proofs}
& \textbf{43}
& \textbf{1}
& \textbf{14} \\
\bottomrule
\end{tabular*}
\caption{
Proof and repair outcomes across three geometry evaluation sets. A statement is
refuted when \textsc{GeoProver} finds a counterexample; final faithful proofs
include both original and human repaired statements.
}
\label{tab:IMO}
\end{table}

We evaluate the complete system on three sets of Olympiad level geometry
problems: 43 historical IMO problems, IMO 2026 Problem~2, and the 14 geometry
statements in \textsc{LEAP}'s Lean-IMO-Bench. Throughout this section, statements are formalized by \textsc{GeoFormalizer} with Claude Opus 4.8 and proofs are constructed by \textsc{GeoProver} with GPT-5.6-Sol. Table~\ref{tab:IMO} reports whether \textsc{GeoProver} proves the original formal statement, refutes it with a Lean verified counterexample, or proves a human repaired version.

\textsc{GeoProver} proves 28 of the 43 statements generated by \textsc{GeoFormalizer}
 within two hours, and one additional statement in 2 hours and
6 minutes. 
Manual inspection confirms that 
all 29 are semantically faithful formalizations of their corresponding
informal problems.
For each of the remaining 14  statements, \textsc{GeoProver} constructs
a formally verified counterexample, typically exposing missing nondegeneracy or
point order assumptions. After expert repair, \textsc{GeoProver} proves all
14 corrected statements, yielding faithful verified proofs for all 43 historical
IMO problems.
\textsc{MechGeo} also formalizes and proves IMO 2026 Problem~2 directly from its informal statement, as illustrated in Figure~\ref{fig:imo2026-case-study}.  On Lean-IMO-Bench, \textsc{GeoProver} proves 12 of the 14 original statements for the first time and refutes the remaining two with counterexamples revealing missing nondegeneracy assumptions. After these two statements are repaired, \textsc{GeoProver} proves both corrected versions.

\subsubsection{Example (IMO 2007 P2, missing nondegeneracy)}
\textbf{$ABCD$ is a parallelogram} and $BCED$ is a cyclic quadrilateral. Let $l$ be a line passing through $A$. Suppose that $l$ intersects the interior of the segment $DC$ at $F$ and intersects line $BC$ at $G$. Suppose also that $EF = EG = EC$. Prove that $l$ is the bisector of $\angle DAB$.

\begin{figure}[h]
    \subfloat[Correct diagram]{\label{fig:imo-nondegenerate}
  \includegraphics[width=0.4\columnwidth]{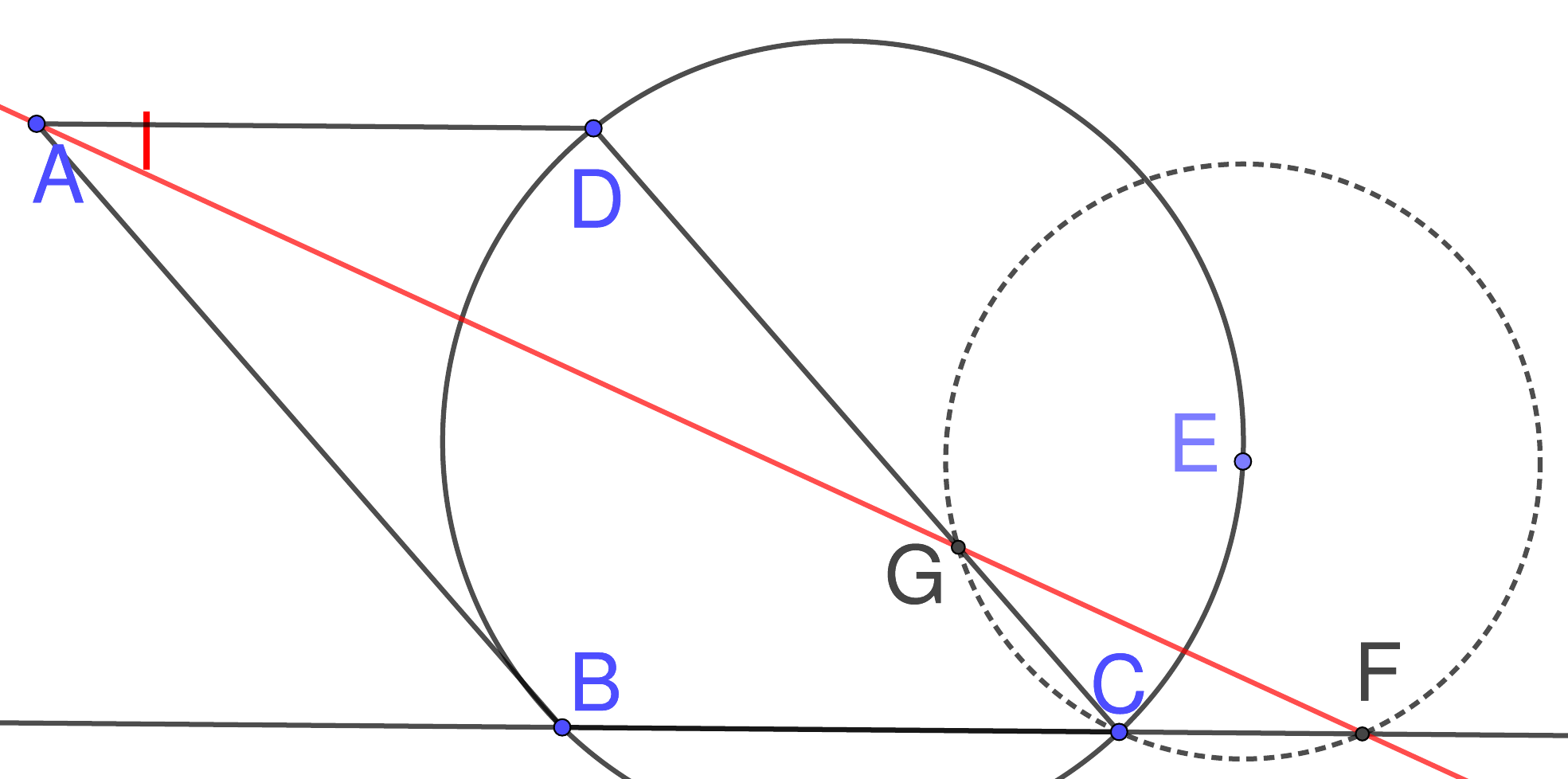}}
    \subfloat[A counterexample to formalization]{
  \includegraphics[width=0.58\columnwidth]{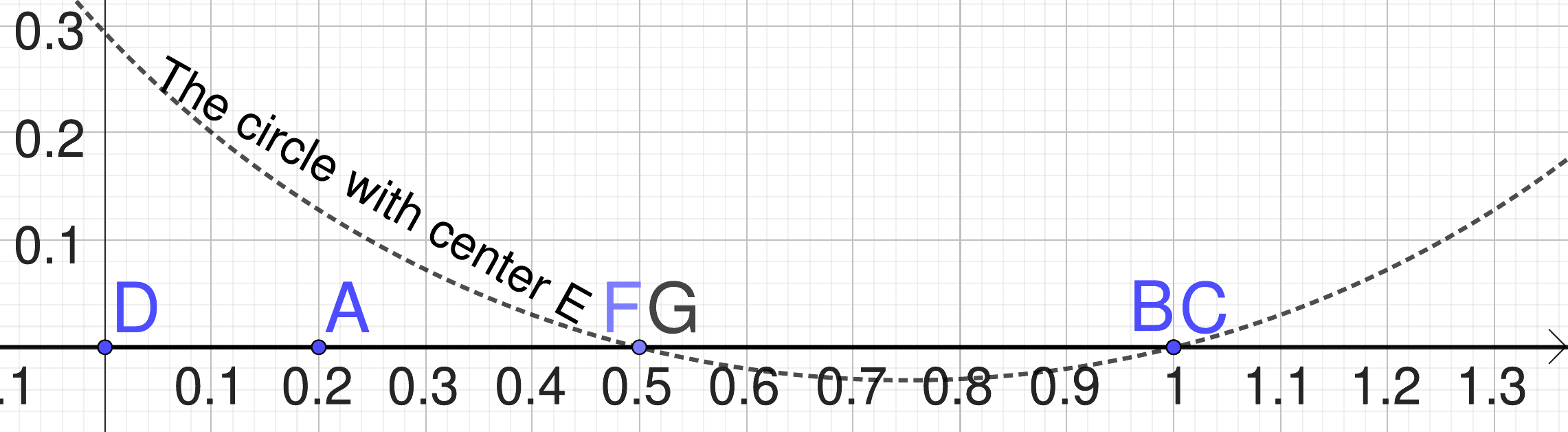}
        \label{fig:imo-degenerate}}
    \caption{IMO 2007 P2}
\end{figure}

Figure~\ref{fig:imo-nondegenerate} shows the intended configuration, while Figure~\ref{fig:imo-degenerate} presents a degenerate counterexample with
$D=(0,0)$, $A=(1/5,0)$, $F=G=(1/2,0)$, $B=C=(1,0)$, $E=(3/4,1)$, for which $\angle DAF=\pi$ and $\angle FAB=0$. The formalization omits the noncollinearity implicit in ``$ABCD$ is a parallelogram.'' After adding that $A$, $B$, and $C$ are noncollinear, \textsc{GeoProver} proves the repaired statement.

\subsubsection{Example (Lean-IMO-Bench  Basic 028, missing order relation)}

In $\triangle ABC$ the altitudes $BE$ and $CF$ intersect at $H$. A circle $(W)$ is
externally tangent to the Euler circle $(E)$ of $\triangle ABC$ and also tangent
to the sides $AB$ and $AC$ at $X$ and $Y$, respectively, with
$(W)$ being closer to $A$ than the Euler circle. Let $I'$ be the
incenter of $\triangle AEF$. Prove that $AXI'Y$ is a rhombus.

\begin{figure}[h]
    \subfloat[Correct diagram]{\label{fig:pt-028}
  \includegraphics[width=0.4\columnwidth]{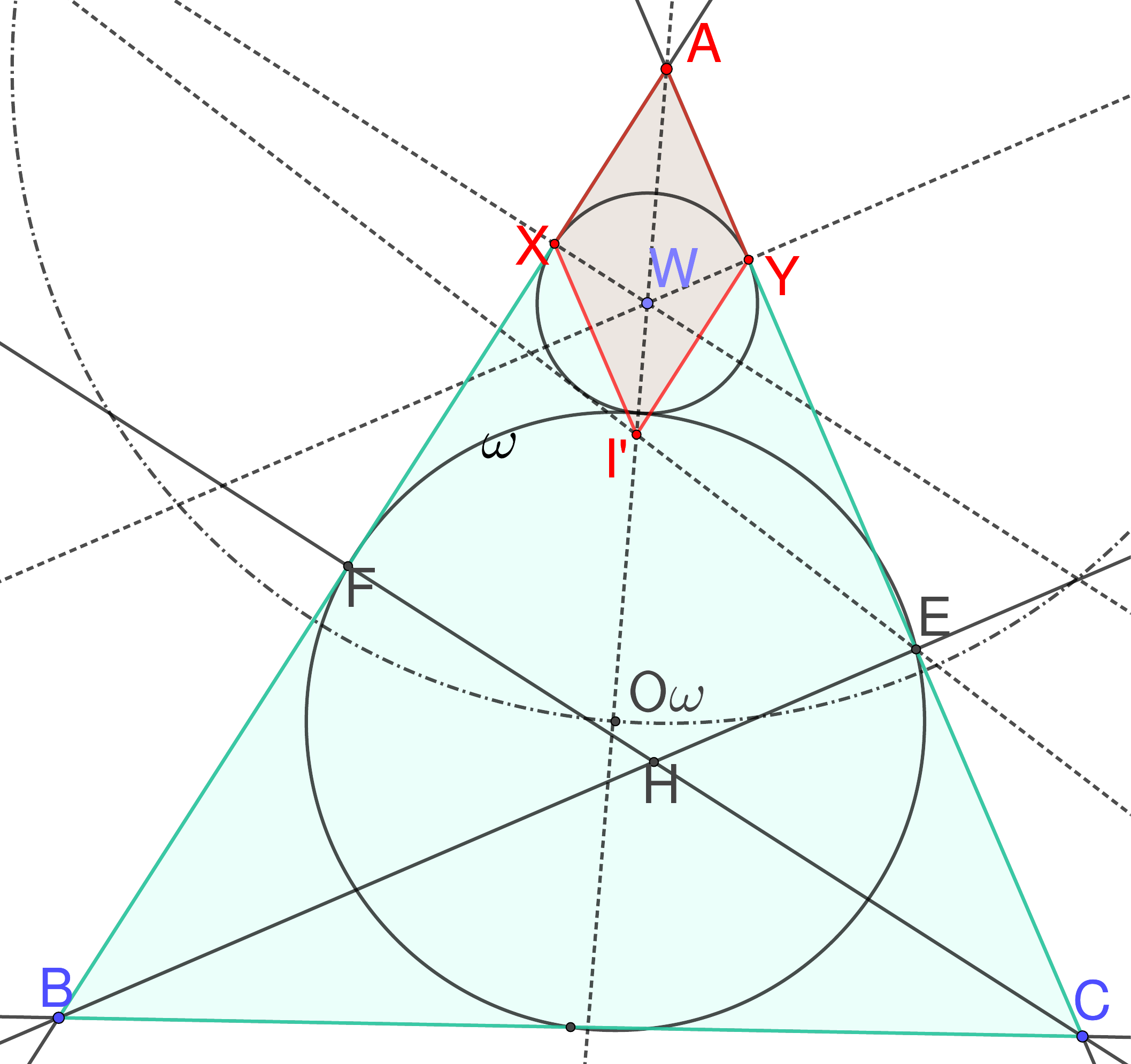}}
    \subfloat[A counterexample to formalization]{
  \includegraphics[width=0.58\columnwidth]{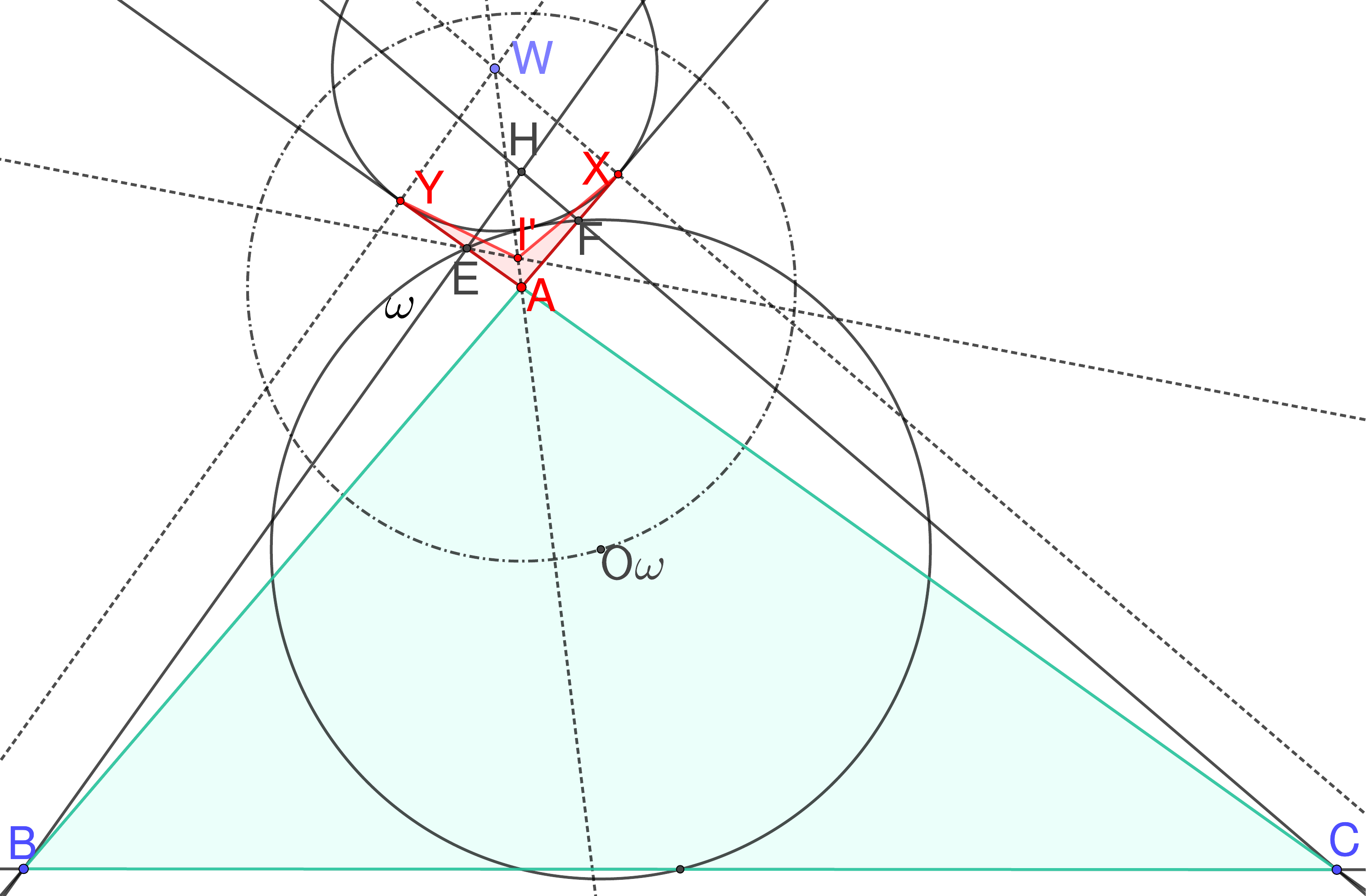}
        \label{fig:pt-028-counterexample}}
    \caption{Lean-IMO-Bench Basic 028}
  \label{fig:pt-028-both}
\end{figure}
Figure~\ref{fig:pt-028-both} 
shows both the intended configuration and an unintended case admitted by the original formalization, in which $X$ and $Y$ lie outside the segments $AB$ and $AC$. After requiring $X$ and $Y$ to lie in the interiors of $AB$ and $AC$, respectively, \textsc{GeoProver} proves the repaired statement.

\section{Conclusion}

Our results show that trustworthy geometry theorem formalization requires
statement validation and proof construction within a unified verification loop.
\textsc{MechGeo} proves challenging geometry theorems, detects incorrect
formalizations through Lean verified counterexamples, and demonstrates the
value of selective algebraization and symbolic computation. Future work will automate semantic and nondegeneracy checks and scale proof
planning and certificate generation to broader and more challenging classes of
geometry theorems.

\section*{Acknowledgment}\label{sec6}

Hao Shen and Lihong Zhi are supported by the National Key R\&D Program of China (Grant No.~2023YFA1009401) and  the Strategic Priority Research Program of the Chinese Academy of Sciences under Grant XDA0480501.  Junyu Guo is encouraged and supported by his supervisor, Xishun Zhao, to work on this project.  We sincerely thank Yunfei Li for helping to review the formalizations of the geometry problems and Ruichen Qiu for assisting with the preparation of Figure~1.

\bibliographystyle{plainnat}
\bibliography{revised}

\setcounter{secnumdepth}{2}
\appendix


\section{Overview of the Appendix}
\label{app:roadmap}

This appendix provides additional technical details and supporting material for \textsc{MechGeo}. We first describe the autoformalization pipeline, including a detailed specification of \textsc{GeoIR}. Next, we present further details of the automated theorem proving process. We then provide several case studies illustrating the effect of proof decomposition on \textsc{GeoProver}'s performance and its ability to discover counterexamples. Finally, we report additional experimental settings.

\section{Autoformalization Details}
\subsection{GeoIR Specification}

Table~3 presents the specification of \textsc{GeoIR} and its deterministic
translation into Mathlib-native Lean. \textsc{GeoIR} represents each problem
as a sequence of object declarations, explicit constructions,
geometric hypotheses, and a single goal.

\subsection{Predicates for Polygon Convexity}
We also introduce two predicates for the formalization of polygon convexity.

\begin{leancode}
def IsConvexQuad (A B C D : EuclideanSpace ℝ (Fin 2)) :=
  letI oangle := (PiLp.basisFun 2 ℝ (Fin 2)).orientation.oangle;
  ((oangle (A -ᵥ B) (C -ᵥ B)).sign = -1 ∧
   (oangle (B -ᵥ C) (D -ᵥ C)).sign = -1 ∧
   (oangle (C -ᵥ D) (A -ᵥ D)).sign = -1 ∧
  (oangle (D -ᵥ A) (B -ᵥ A)).sign = -1) ∨
  ((oangle (A -ᵥ B) (C -ᵥ B)).sign = 1 ∧
   (oangle (B -ᵥ C) (D -ᵥ C)).sign = 1 ∧
   (oangle (C -ᵥ D) (A -ᵥ D)).sign = 1 ∧
   (oangle (D -ᵥ A) (B -ᵥ A)).sign = 1)

def IsConvex {n} (f : Fin n → EuclideanSpace ℝ (Fin 2)) : Prop :=
  letI oangle := (PiLp.basisFun 2 ℝ (Fin 2)).orientation.oangle;
  (∀ i j : Fin n,
    j ≠ i → j ≠ finRotate n i →
      (oangle (f (finRotate n i) -ᵥ f i) (f j -ᵥ f i)).sign = 1) ∨
  (∀ i j : Fin n,
    j ≠ i → j ≠ finRotate n i →
      (oangle (f (finRotate n i) -ᵥ f i) (f j -ᵥ f i)).sign = -1)
\end{leancode}

\section{Automated Theorem Proving Details}

\subsection{Algebraization Lemmas}

The conversion of geometric objects and expressions into algebraic form, primarily polynomial equalities and inequalities, proceeds in two steps. Each step can be performed either by applying the lemmas listed in the docstring of \texttt{MechGeoBench/GeoTheorem/ToPolyMacros.lean} or by using the following tactics:

\begin{enumerate}
  \item convert derived expressions to basic expressions: the ``Derived'' part, the \lean{to_basic} tactic;
  \item convert basic expressions to algebraic expressions: the ``Basic''  part, the \lean{basic_to_poly} tactic.
\end{enumerate}
The two steps can also be done in a single tactic \lean{to_poly}. These three tactics can be called with an extra argument \lean{at POSITION(s)} like \lean{simp}, such as \lean{to_poly at h}, which is for selective algebraization.

Step 1 and its reverse can also serve as a bridge between different derived expressions.

\subsection{Proof Efficiency}
\begin{figure}[htbp]
      \centering
      \includegraphics[width=0.6\linewidth]{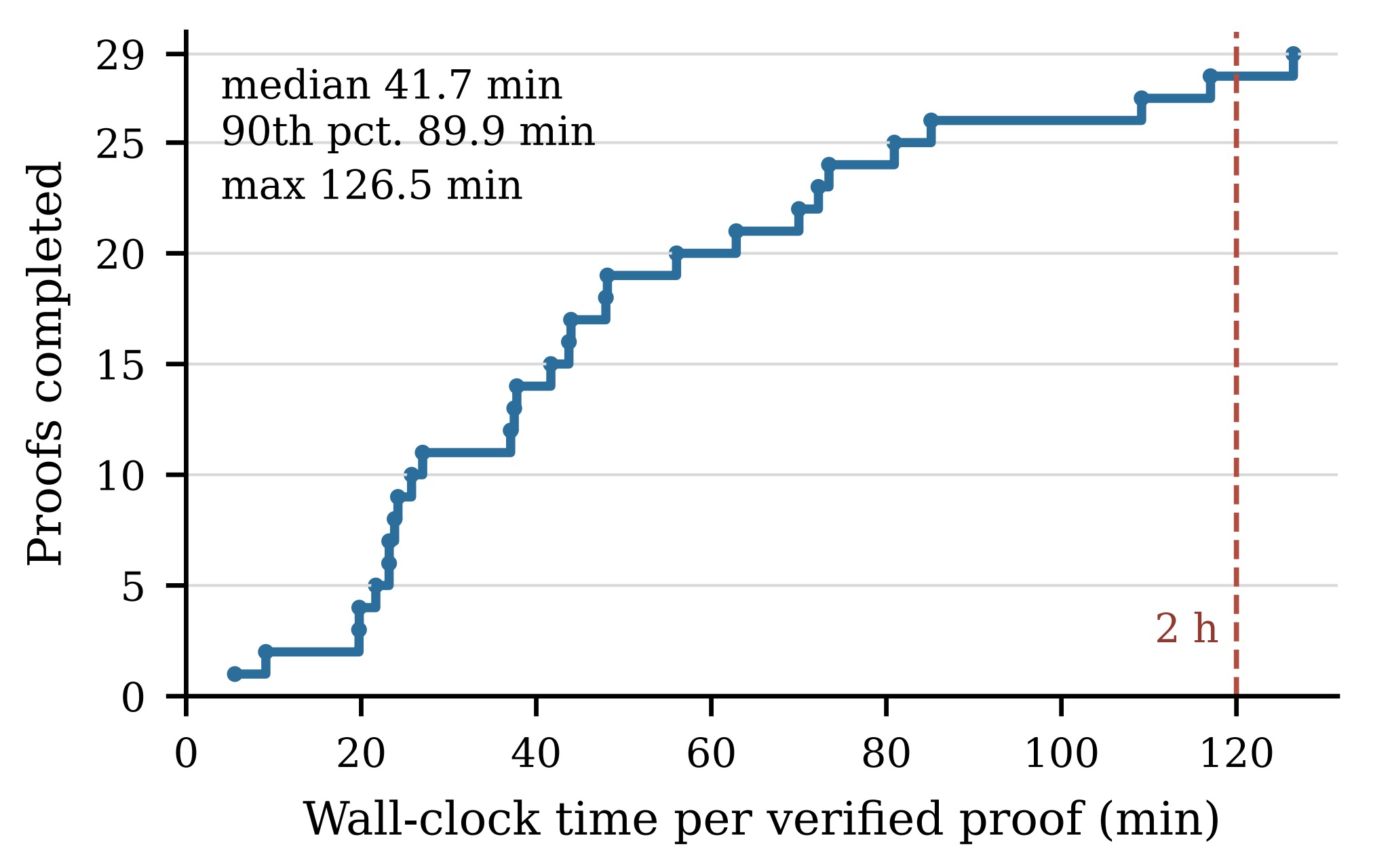}
      \caption{Cumulative distribution of wall-clock time for the 29 proofs of the original formalizations. The dashed vertical line marks the two-hour reference point
corresponding to the budget used in the main comparison experiments.} 
      \label{fig:imo-29}
\end{figure}

\begin{figure}[htbp]
      \centering
      \includegraphics[width=0.6\linewidth]{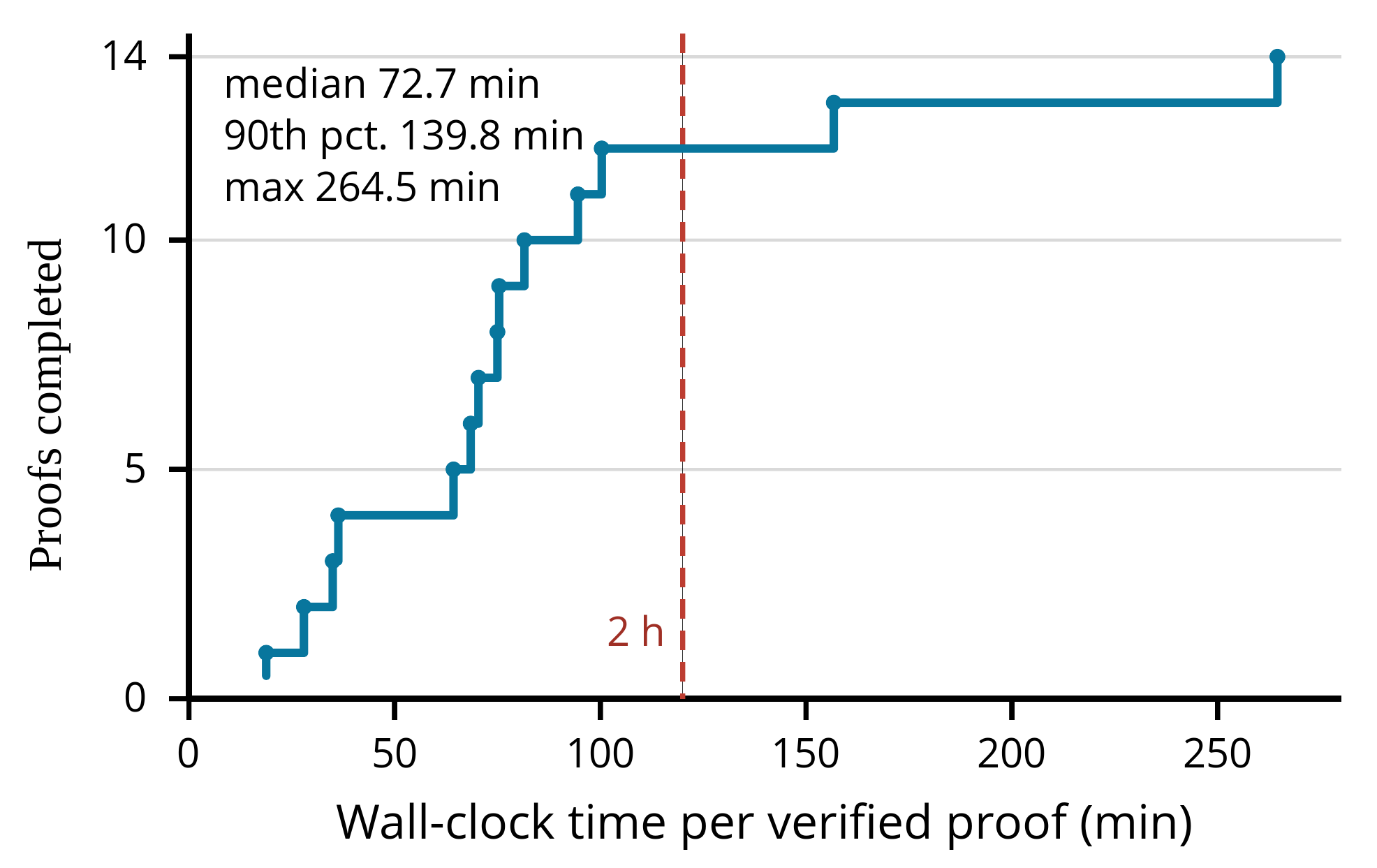}
      \caption{Cumulative distribution of wall-clock time for the 14 proofs of the expert-repaired formalizations. The dashed vertical line marks the two-hour reference point corresponding to the budget used in the main comparison experiments. 
}
      \label{fig:imo-14}
\end{figure}

 To investigate the additional proof coverage obtainable under an extended budget by \textsc{GeoProver}, we subsequently conduct a separate run using GPT-5.6-Sol, with a timeout of 24 hours per problem.  
The dashed vertical line marks the two-hour reference point
corresponding to the budget used in the main comparison experiments.
Figure~\ref{fig:imo-29} covers the 29 formalizations generated by \textsc{GeoFormalizer} using Claude Opus~4.8 proved directly by \textsc{GeoProver} without human repair. Figure~\ref{fig:imo-14} reports the corresponding results for the 14 expert-repaired formalizations.

\subsection{Detailed Results on Lean-IMO-Bench}
Table~\ref{tab:rq3-leap} compares proof success on the 14 original
geometry statements in LEAP's Lean-IMO-Bench. \textsc{MechGeo} obtains kernel-checked
proofs for 12 of the 14 original statements in total. For the remaining two statements, \textsc{GeoProver}
constructs Lean-verified counterexamples that expose missing assumptions.
After experts repair these two statements, \textsc{MechGeo} proves both
corrected versions, yielding faithful verified proofs for all 14 problems.
Because the repaired statements differ from the original benchmark inputs,
we retain 12/14 as the directly comparable leaderboard result and report
14/14 separately as the final coverage after validation and repair. (See \texttt{MechGeoBench/LEAP/} in the supplementary material for Lean source code.)

\begin{table}[t]
\centering
\small
\setlength{\tabcolsep}{4pt}
\begin{tabular*}{\columnwidth}{@{\extracolsep{\fill}}lccc@{}}
\toprule
Method & Basic (6) & Advanced (8) & Total (14) \\
\midrule
Hilbert          & 0 & 0 & 0 (0.0\%) \\
Gemini 3.1 Pro   & 0 & 0 & 0 (0.0\%) \\
Goedel-V2-32B    & 0 & 0 & 0 (0.0\%) \\
Aristotle        & 1 & 0 & 1 (7.1\%) \\
LEAP             & 1 & 1 & 2 (14.3\%) \\
\textsc{MechGeo} & \textbf{5} & \textbf{7} &
                    \textbf{12 (85.7\%)} \\
\bottomrule
\end{tabular*}
\caption{Verified proofs on the 14 original geometry statements in
LEAP's Lean-IMO-Bench. Baseline results are taken
from the official leaderboard.}
\label{tab:rq3-leap}
\end{table}

\section{Case Studies}
\subsection{IMO 2008 P1}
\begin{statement}
An acute-angled triangle $ABC$ has orthocentre $H$. The circle passing
through $H$ with centre the midpoint of $BC$ intersects the line $BC$ at
$A_1$ and $A_2$. Similarly, the circle passing through $H$ with centre the
midpoint of $CA$ intersects the line $CA$ at $B_1$ and $B_2$, and the circle
passing through $H$ with centre the midpoint of $AB$ intersects the line $AB$
at $C_1$ and $C_2$. Show that $A_1$, $A_2$, $B_1$, $B_2$, $C_1$, and $C_2$
lie on a circle. 
\end{statement}

\begin{figure}[h]
\centering
  \includegraphics[width=0.6\linewidth]{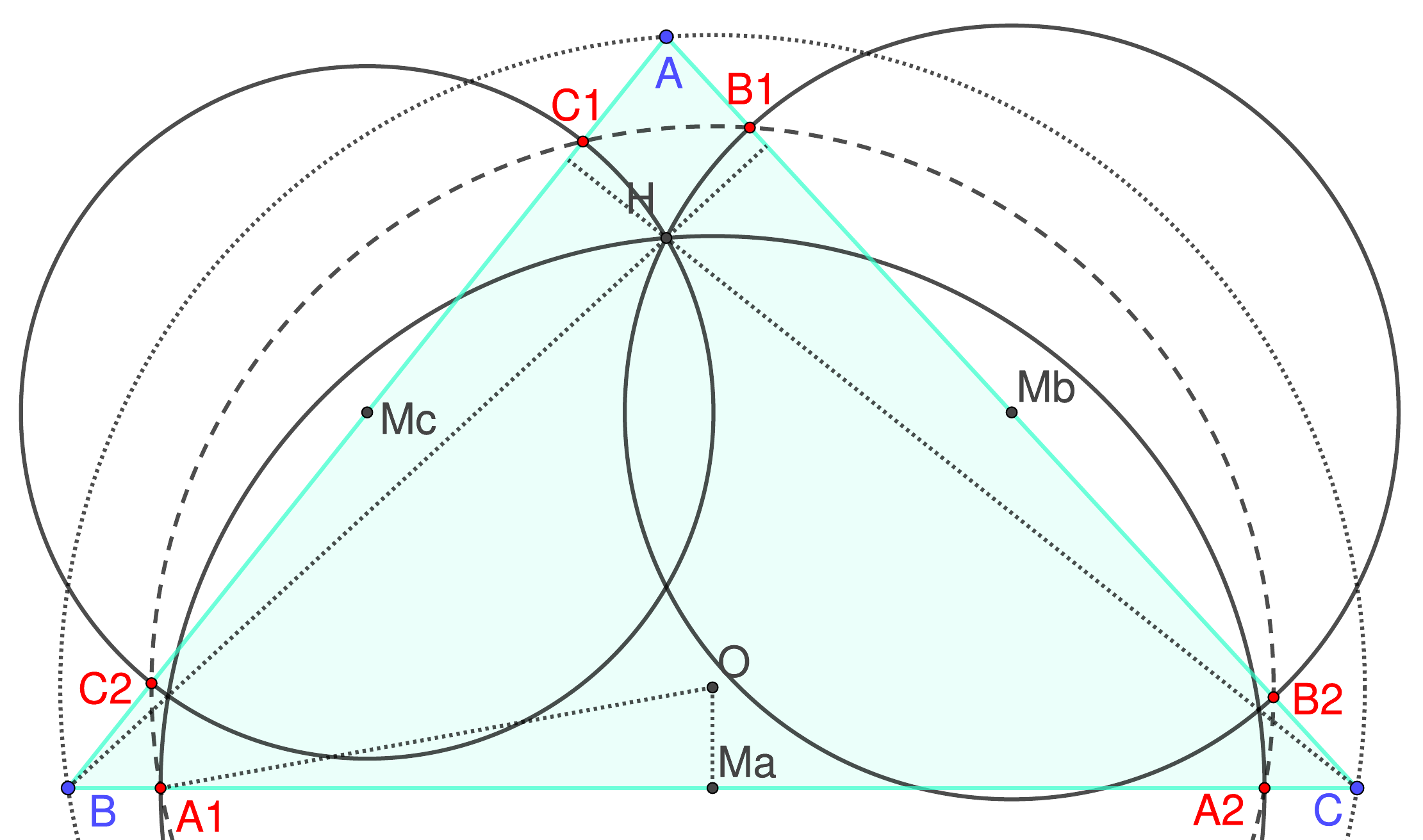}
  \caption{IMO 2008 P1}
  \label{fig:imo-2008-p1-appendix}
\end{figure}

\paragraph{Why global algebraization is difficult.}
As shown in Figure~\ref{fig:imo-2008-p1-appendix}, if there is such a circle, one can show that  it has the same center as the circumcircle of $\triangle ABC$. Denote this center by $O$, then it suffices to prove $OA_1=OA_2=OB_1=OB_2=OC_1=OC_2$. A direct coordinate encoding introduces two scalar coordinates for each of 
\[
 A,B,C,H,A_1,A_2,B_1,B_2,C_1,C_2,O,
\]
and therefore contains 22 scalar variables.  Its 16 quadratic
equations consist of two altitude constraints, six collinearity constraints,
six midpoint-circle constraints, and two circumcentre constraints.
Algebraizing these constraints and the conclusion as a single system creates
an unnecessarily large elimination problem.

\paragraph{Proof Decomposition.}
Instead of translating the entire theorem into one large
polynomial system, \textsc{GeoProver} first constructs a proof plan. Let \(M_a\), \(M_b\), and \(M_c\) denote the midpoints of
\(BC\), \(CA\), and \(AB\), respectively.
The proof is decomposed into the following obligations:
\[
OA_1^2 = OA_2^2
       = OM_a^2 + HM_a^2,
\]
\[
OB_1^2 = OB_2^2
       = OM_b^2 + HM_b^2,
\]
\[
OC_1^2 = OC_2^2
       = OM_c^2 + HM_c^2,
\]
together with
\[
OM_a^2 + HM_a^2
=
OM_b^2 + HM_b^2
=
OM_c^2 + HM_c^2.
\]

These identities immediately imply
\[
OA_1=OA_2=OB_1=OB_2=OC_1=OC_2,
\]
and therefore the six points lie on a common circle centered
at \(O\).

Rather than solving one monolithic elimination problem
containing all 22 scalar variables and 16 quadratic
constraints, \textsc{GeoProver} reduces the proof to several
independent polynomial obligations, each involving at most ten scalar variables.  The remainder of this section uses the first obligation,
\[
OA_1^2 = OM_a^2 + HM_a^2,
\]
as a representative example. For the point \(A_1\), the local system only involves the
coordinates of \(A_1\), \(B\), \(C\), \(H\), and \(O\).
Since \(M_a\) is uniquely determined by \(B\) and \(C\), the
system contains at most ten  scalar coordinate variables.
Its constraints are simply $OB=OC$, $B,C,A_1\text{ are collinear}$, and $
M_aA_1=HM_a$,
together with the nondegeneracy condition \(B\neq C\).

Because \(O\) is the circumcenter,
\(OM_a\perp BC\).
Since \(A_1\), \(B\), and \(C\) are collinear,
\(A_1M_a\parallel BC\), and therefore
\(A_1M_a\perp OM_a\).
Hence
\[
OA_1^2
=
OM_a^2
+
M_aA_1^2.
\]
Using the circle constraint
\(M_aA_1=HM_a\),
GeoProver derives
\[
OA_1^2
=
OM_a^2
+
HM_a^2.
\]
The same argument is then instantiated for
\(A_2\), \(B_1\), \(B_2\), \(C_1\), and \(C_2\). The corresponding code is as follows:
\begin{leancode}
have h_OA1_sq : Inner.inner ℝ (A1 - O) (A1 - O) = Inner.inner ℝ (Ma - O) (Ma - O) + Inner.inner ℝ (H - Ma) (H - Ma) := by
    calc
      Inner.inner ℝ (A1 - O) (A1 - O) =
          Inner.inner ℝ (A1 - Ma) (A1 - Ma) + Inner.inner ℝ (Ma - O) (Ma - O) +
          2 * Inner.inner ℝ (A1 - Ma) (Ma - O) := by rw [inner_sq_expand A1 Ma O]
      _ = Inner.inner ℝ (H - Ma) (H - Ma) + Inner.inner ℝ (Ma - O) (Ma - O) +
          2 * Inner.inner ℝ (A1 - Ma) (Ma - O) := by rw [h_distA1_sq]
      _ = Inner.inner ℝ (H - Ma) (H - Ma) + Inner.inner ℝ (Ma - O) (Ma - O) +
          2 * (0 : ℝ) := by rw [h_inner_A1_zero]
      _ = Inner.inner ℝ (Ma - O) (Ma - O) + Inner.inner ℝ (H - Ma) (H - Ma) := by ring
\end{leancode}


\subsection{IMO 2007 P2}
\paragraph{Informal Statement}
\begin{statement}
Consider five points $A, B, C, D$ and $E$ such that $ABCD$ is a parallelogram and $BCED$ is a cyclic quadrilateral. Let $l$ be a line passing through $A$. Suppose that $l$ intersects the interior of the segment $DC$ at $F$ and intersects line $BC$ at $G$. Suppose also that $EF = EG = EC$. Prove that $l$ is the bisector of angle $DAB$.
\end{statement}

\paragraph{Formal Statement} (generated by \textsc{GeoFormalizer} and \textcolor{leanrepaircolor}{repaired manually})
\leavevmode

\begin{leancode}
theorem IMOBench_IMO_2007_p2 :
    ∀ A : EuclideanSpace ℝ (Fin 2),
    ∀ B : EuclideanSpace ℝ (Fin 2),
    ∀ C : EuclideanSpace ℝ (Fin 2),
    ∀ D : EuclideanSpace ℝ (Fin 2),
    ∀ E : EuclideanSpace ℝ (Fin 2),
    ∀ F : EuclideanSpace ℝ (Fin 2),
    ∀ G : EuclideanSpace ℝ (Fin 2),
    (¬ LinearIndependent (M := EuclideanSpace ℝ (Fin 2)) ℝ ![A -ᵥ B, C -ᵥ D] ∧ ¬ LinearIndependent (M := EuclideanSpace ℝ (Fin 2)) ℝ ![A -ᵥ D, C -ᵥ B]) →
(*\beginleanrepair*)    AffineIndependent ℝ ![A, B, D] →(*\endleanrepair*)
    EuclideanGeometry.Concyclic (P := EuclideanSpace ℝ (Fin 2)) {B, C, E, D} →
    Sbtw ℝ D F C →
    Collinear ℝ {A, F, G} →
    Collinear ℝ {B, C, G} →
    (Dist.dist (α := EuclideanSpace ℝ (Fin 2)) E F) = (Dist.dist (α := EuclideanSpace ℝ (Fin 2)) E C) →
    (Dist.dist (α := EuclideanSpace ℝ (Fin 2)) E G) = (Dist.dist (α := EuclideanSpace ℝ (Fin 2)) E C) →
    (D ≠ A ∧ A ≠ B) →
    ((EuclideanGeometry.angle (P := EuclideanSpace ℝ (Fin 2)) D A F) = (EuclideanGeometry.angle (P := EuclideanSpace ℝ (Fin 2)) F A B)) := by
  sorry
\end{leancode}

\paragraph{Counterexample and Repair}

\begin{figure}[h]
    \centering
    \includegraphics[width=0.6\linewidth]{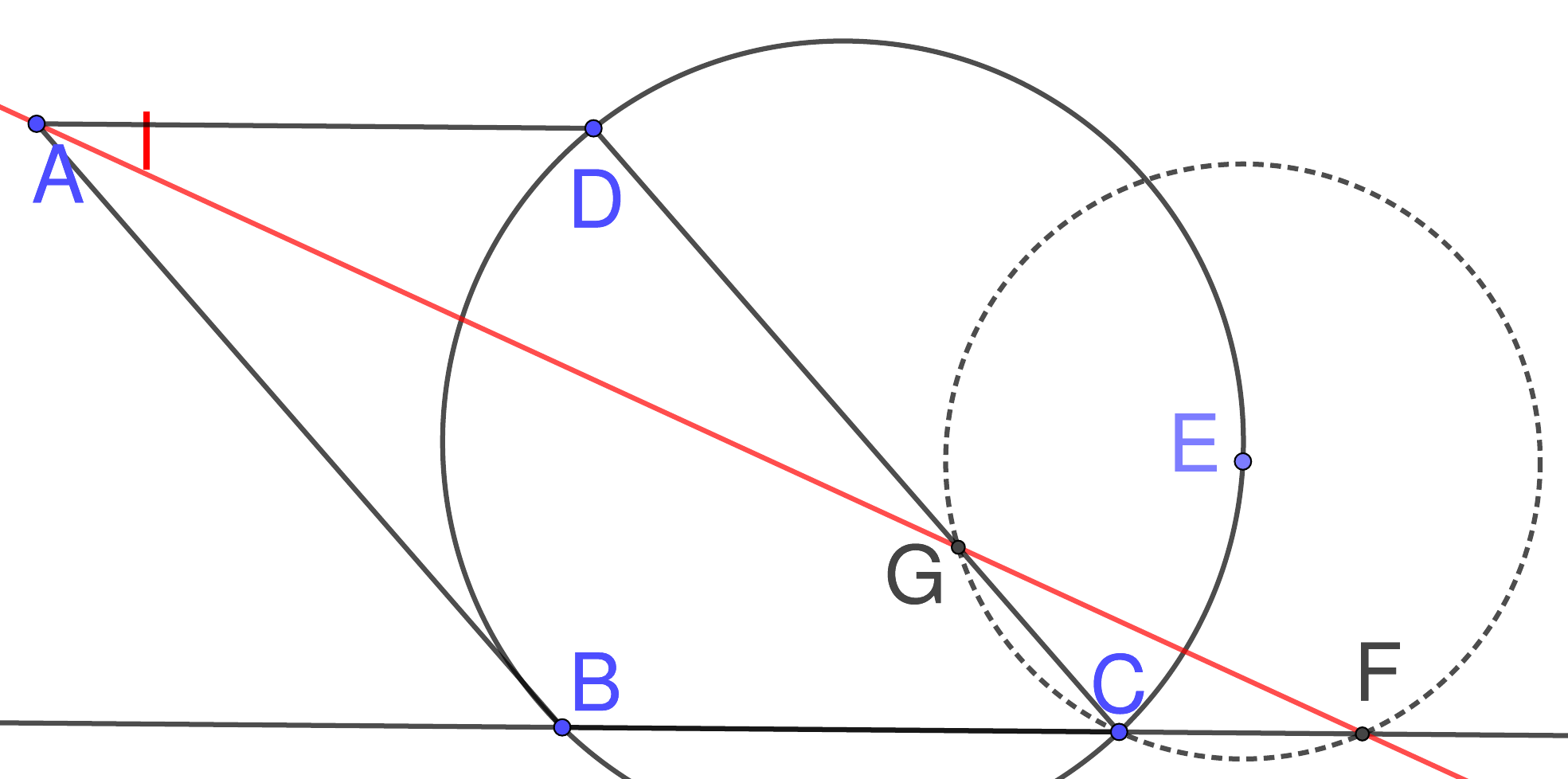}
    \caption{The intended configuration of IMO 2007 P2.}
    \label{fig:imo-2027-appendix}
\end{figure}

\begin{figure}[h]
   \centering
    \includegraphics[width=0.6\linewidth]{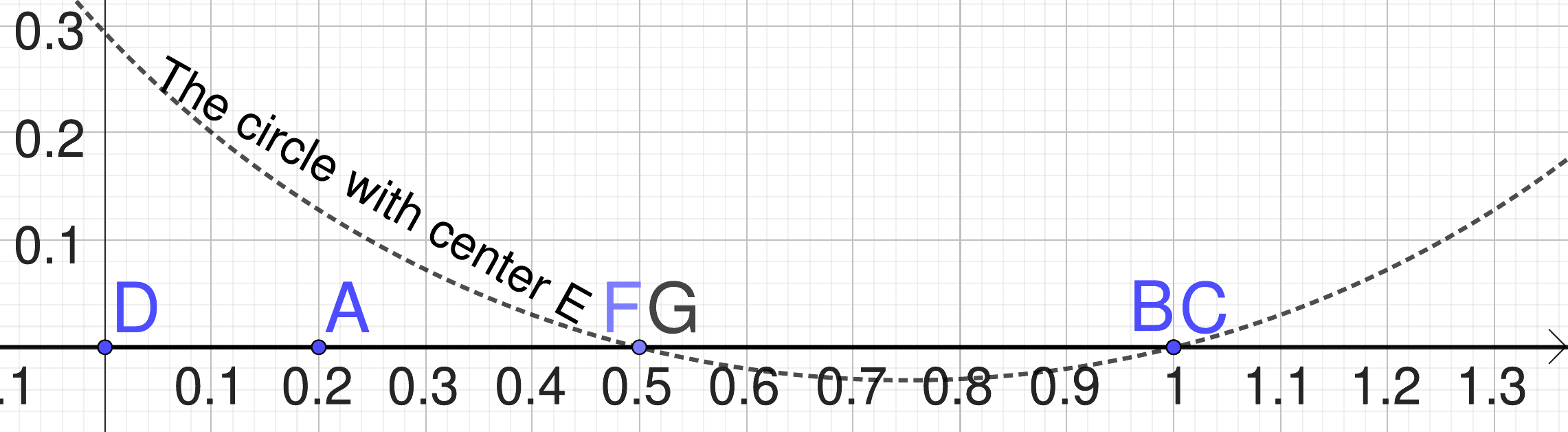}
    \caption{An unintended counterexample found by \textsc{GeoProver} in the generated formalization of IMO 2007 P2, where $ABCD$ is degenerate.}
    \label{fig:imo-2027-degenerate-appendix}
\end{figure}

In the generated formalization, the condition ``$ABCD$ is a parallelogram'' is formalized as ``$\overrightarrow{AB}$ and $\overrightarrow{CD}$, as well as $\overrightarrow{AD}$ and $\overrightarrow{CB}$, are linearly dependent'', i.e. \lean{¬ LinearIndependent ℝ ![A -ᵥ B, C -ᵥ D] ∧ ¬ LinearIndependent ℝ ![A -ᵥ D, C -ᵥ B])}.
The condition ``$BCED$ is a cyclic quadrilateral'' is formalized as ``$BCED$ is concyclic.'' The goal is formalized as $\angle DAF = \angle FAB$.

Figure~\ref{fig:imo-2027-appendix} shows the intended configuration, while Figure~\ref{fig:imo-2027-degenerate-appendix} presents a degenerate counterexample with
$D=(0,0)$, $A=(1/5,0)$, $F=G=(1/2,0)$, $B=C=(1,0)$, $E=(3/4,1)$, where the parallelogram $ABCD$ is degenerate to a segment, for which $\angle DAF=\pi$ and $\angle FAB=0$. The formalization omits the noncollinearity implicit in ``$ABCD$ is a parallelogram.'' After adding that $A$, $B$, and $D$ are noncollinear (\lean{AffineIndependent ℝ ![A, B, D]}, the green line), \textsc{GeoProver} proves the repaired statement.

\subsection{Lean-IMO-Bench Basic 028}

\paragraph{Informal Statement}

\begin{statement}
In $\triangle ABC$ the altitudes $BE$ and $CF$ intersect at $H$. A circle $(W)$ is
externally tangent to the Euler circle $(E)$ of $\triangle ABC$ and also tangent
to the sides $AB$ and $AC$ at $X$ and $Y$, respectively, with
$(W)$ being closer to $A$ than the Euler circle. Let $I'$ be the
incenter of $\triangle AEF$. Prove that $AXI'Y$ is a rhombus.
\end{statement}

\paragraph{Formal Statement} (from LEAP and \textcolor{leanrepaircolor}{repaired manually})

\begin{leancode}
local notation "ℝ²" => EuclideanSpace ℝ (Fin 2)

theorem PBBasic028
    (A B C : ℝ²) (tri : AffineIndependent ℝ ![A, B, C])
    (ω : Sphere ℝ²)
    (hω : midpoint ℝ A B ∈ ω ∧ midpoint ℝ B C ∈ ω ∧ midpoint ℝ C A ∈ ω)
    (W : Sphere ℝ²) (X Y : ℝ²)
    (hX : W.IsTangentAt X (affineSpan ℝ {A, B}))
    (hY : W.IsTangentAt Y (affineSpan ℝ {A, C}))
(*\beginleanrepair*)    -- The tangency points lie on the actual sides, not their extended lines.
    (hXseg : Wbtw ℝ A X B)
    (hYseg : Wbtw ℝ A Y C)(*\endleanrepair*)
    (tangent : W.IsExtTangent ω)
    (closer : dist A W.center < dist A ω.center) :
    let E := altitudeFoot ⟨![A, B, C], tri⟩ 1
    let F := altitudeFoot ⟨![A, B, C], tri⟩ 2
    let H := Triangle.orthocenter ⟨![A, B, C], tri⟩
    ∀ tri2 : AffineIndependent ℝ ![A, E, F],
    let I' := incenter ⟨![A, E, F], tri2⟩
    List.Pairwise (· = ·) [dist A X, dist X I', dist I' Y, dist Y A]
\end{leancode}

\paragraph{Counterexample and Repair}

\begin{figure}[h]
\centering
\includegraphics[width=0.6\linewidth]{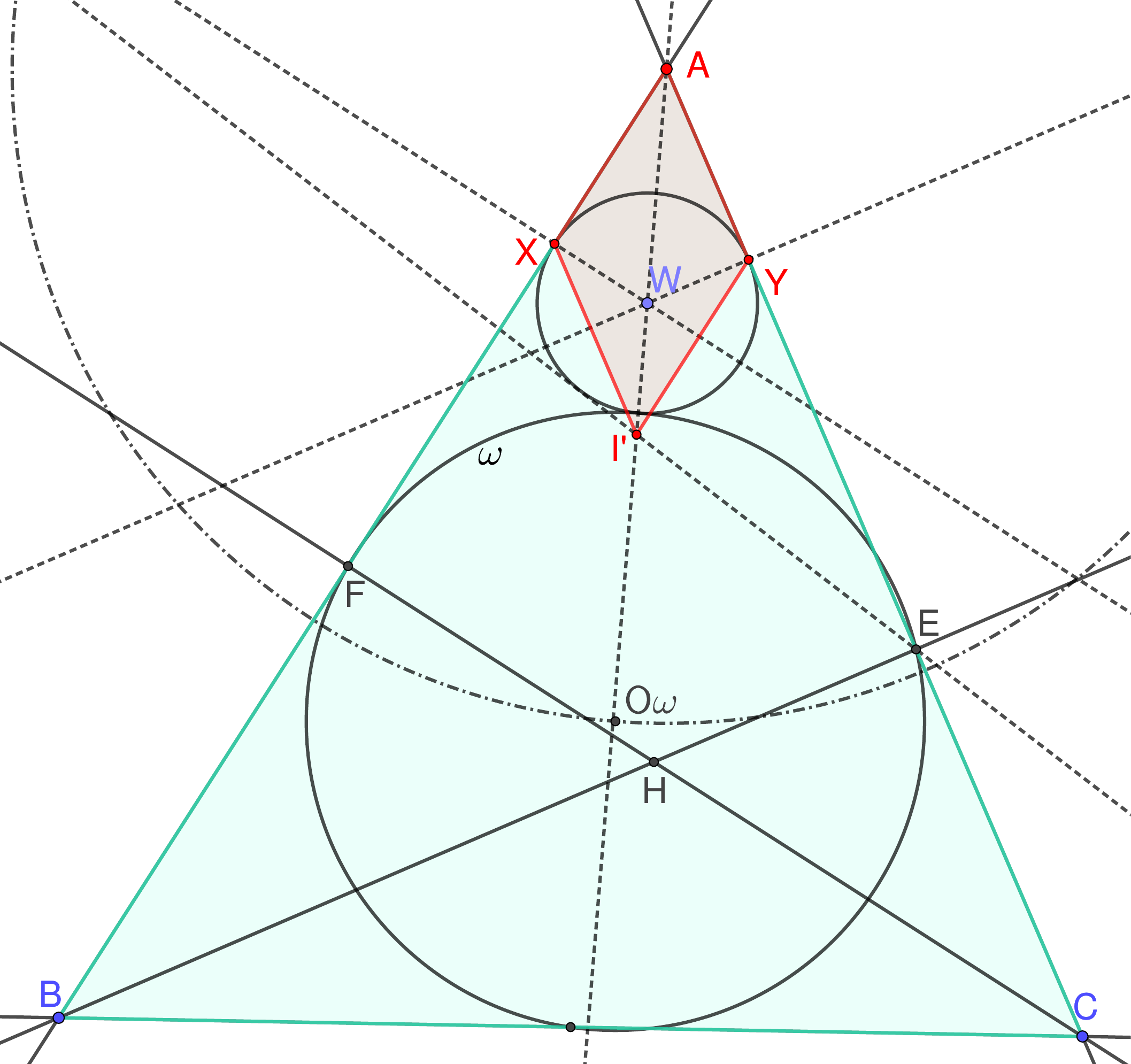}
\caption{The intended configuration of Lean-IMO-Bench Basic 028}
\label{fig:leap-028-appendix}
\end{figure}

\begin{figure}[h]
\centering
\includegraphics[width=0.6\linewidth]{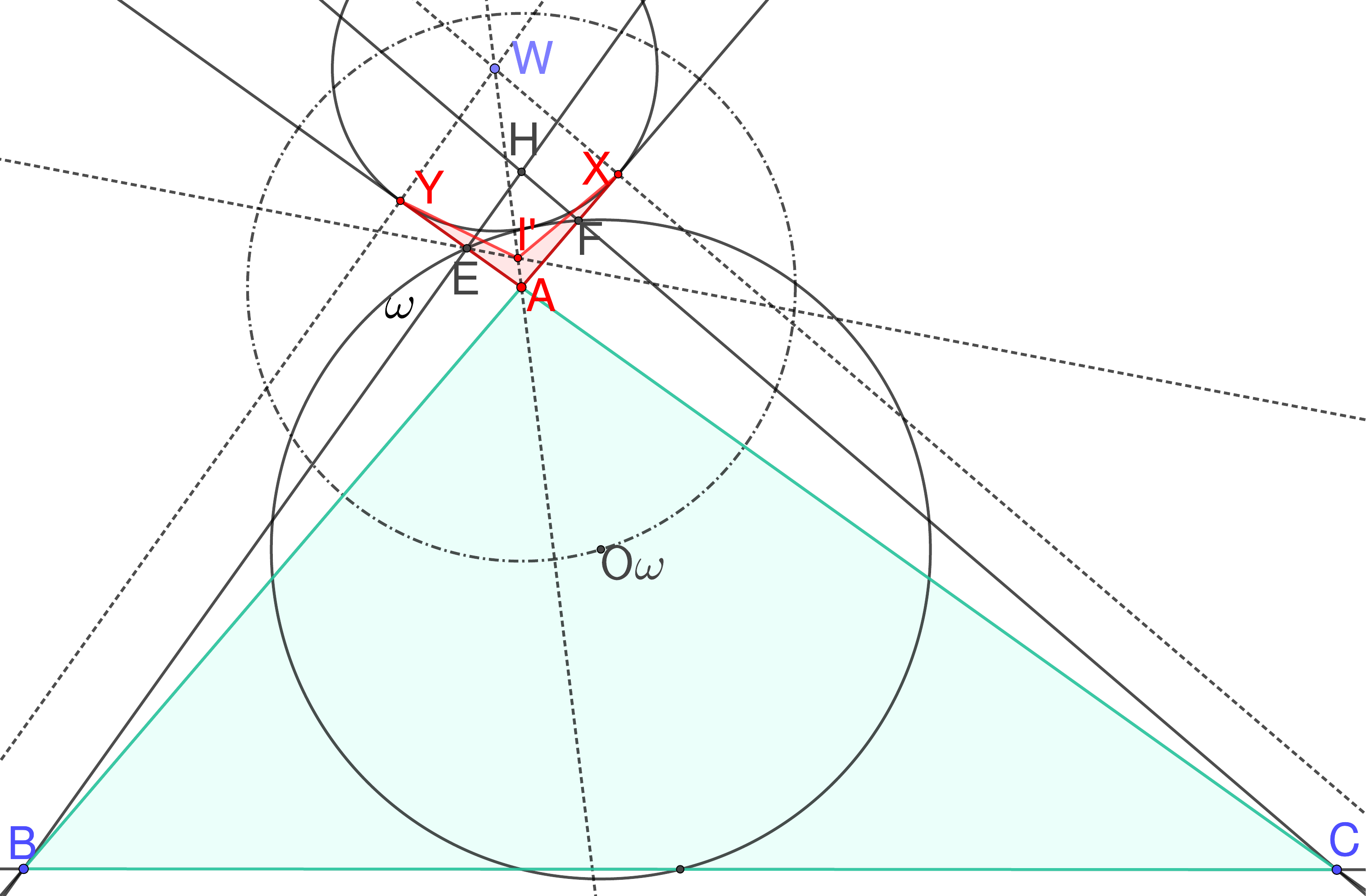}
\caption{Unintended case allowed by the original formalization of Lean-IMO-Bench Basic 028, where $X$ and $Y$ are outside of the segments $AB$ and $AC$ and thus $AXI'Y$ is concave.}
\label{fig:leap-028-counter-appendix}
\end{figure}

The intended configuration is shown in Figure~\ref{fig:leap-028-appendix}. However, the original formalization does not require $X$ and $Y$ to lie on the segments $AB$ and $AC$, respectively, although these conditions are  necessary for the conclusion to hold. If $X$ and $Y$ lie on the extended lines outside these segments, the quadrilateral $AXI'Y$ may become concave, as illustrated in Figure~\ref{fig:leap-028-counter-appendix}. We therefore add \lean{Wbtw ℝ A X B} and \lean{Wbtw ℝ A Y C}, as required in the informal statement that the tangency points lie on the actual sides. With these assumptions,
\textsc{GeoProver} proves the repaired statement.

\subsection{Lean-IMO-Bench Advanced 010}
\paragraph{Informal Statement.}
\begin{statement}
Let \(O\) and \(G\) be the circumcenter and centroid of a non-isosceles
triangle \(ABC\), respectively. Let \(H\) be the foot of the perpendicular
from \(A\) to \(BC\), and let \(M\) be the midpoint of \(BC\).

For a point \(X\) on the line \(OG\), let the line \(BX\) intersect \(AC\)
at \(P\), and let the line \(CX\) intersect \(AB\) at \(Q\). Let \(H_1\) be
the foot of the perpendicular from \(P\) to the line \(AB\), and let \(K\)
be the reflection of \(A\) about \(H_1\).

Let \(T\) be the intersection other than \(P\) of the circumcircle of
triangle \(KPQ\) and the circumcircle of triangle \(PHM\).

Prove that, as \(X\) moves along the line \(OG\), the point \(T\) moves
along a fixed circle.
\end{statement}

\subsubsection{Formal Statement} (from LEAP and \textcolor{leanrepaircolor}{repaired manually})
\leavevmode\par
\begin{leancode}
local notation "ℝ²" => EuclideanSpace ℝ (Fin 2)

theorem PBAdvanced010
    -- a triangle ABC
    (A B C : ℝ²) (tri : AffineIndependent ℝ ![A, B, C])
    -- ABC is non-isosceles
    (n_isosceles : [dist A B, dist A C, dist B C].Nodup)
    (O : ℝ²) (hO : O = circumcenter ⟨![A, B, C], tri⟩)
    (G : ℝ²) (hG : G = Finset.centroid ℝ .univ ![A, B, C])
    (H : ℝ²) (hH : H = altitudeFoot ⟨![A, B, C], tri⟩ 0)
    (M : ℝ²) (hM : M = midpoint ℝ B C) :
    ∃ ω : Sphere ℝ²,
    ∀ (X : ℝ²) (hX : Collinear ℝ {X, O, G})
    (P : ℝ²) (hP : Collinear ℝ {P, B, X} ∧ Collinear ℝ {P, A, C})
    (Q : ℝ²) (hQ : Collinear ℝ {Q, C, X} ∧ Collinear ℝ {Q, A, B})
    (H₁ : ℝ²) (hH₁ : H₁ = orthogonalProjection (affineSpan ℝ {A, B}) P)
    (K : ℝ²) (hK : K = reflection (affineSpan ℝ {H₁}) A)
(*\beginleanrepair*)    -- Non-degeneracy of two triangles
    (hKPQ : AffineIndependent ℝ ![K, P, Q])
    (hPHM : AffineIndependent ℝ ![P, H, M])
    -- Circumcircles of them are different
    (hcircles :
      circumsphere (⟨![K, P, Q], hKPQ⟩ : Affine.Simplex ℝ ℝ² 2) ≠
      circumsphere (⟨![P, H, M], hPHM⟩ : Affine.Simplex ℝ ℝ² 2))(*\endleanrepair*)
    (T : ℝ²) (hT : T ≠ P ∧ Cospherical {T, K, P, Q} ∧ Cospherical {T, P, H, M}),
    T ∈ ω := by sorry
\end{leancode}

\begin{figure}[h]
    \centering
    \includegraphics[width=0.6\linewidth]{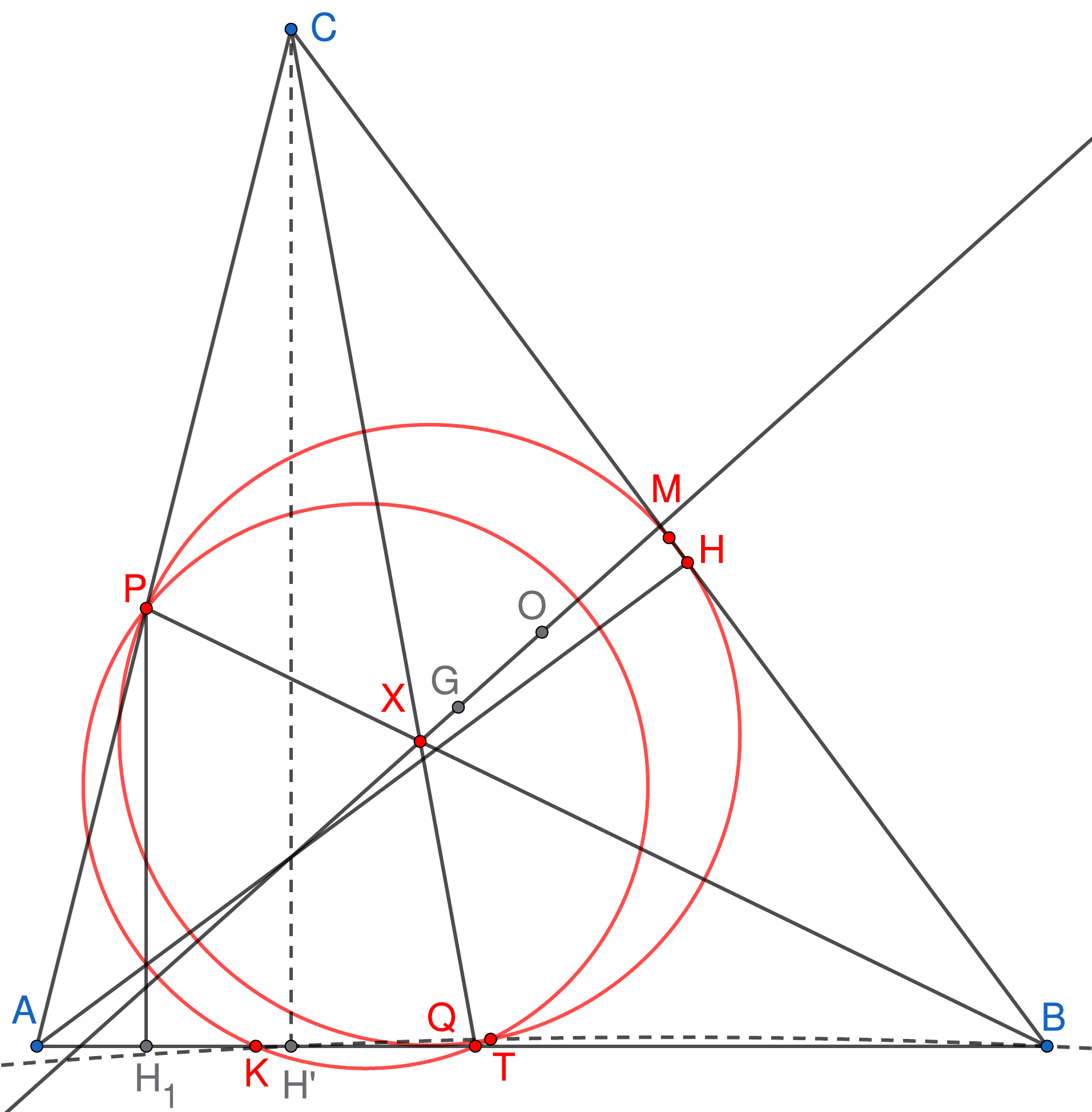}
    \caption{The intended configuration of Lean-IMO-Bench Advanced 010, where the dashed curve through \(T\) is the locus of $T$}
    \label{fig:010-appendix}
\end{figure}

\begin{figure}[h]
    \centering
    \includegraphics[width=0.6\linewidth]{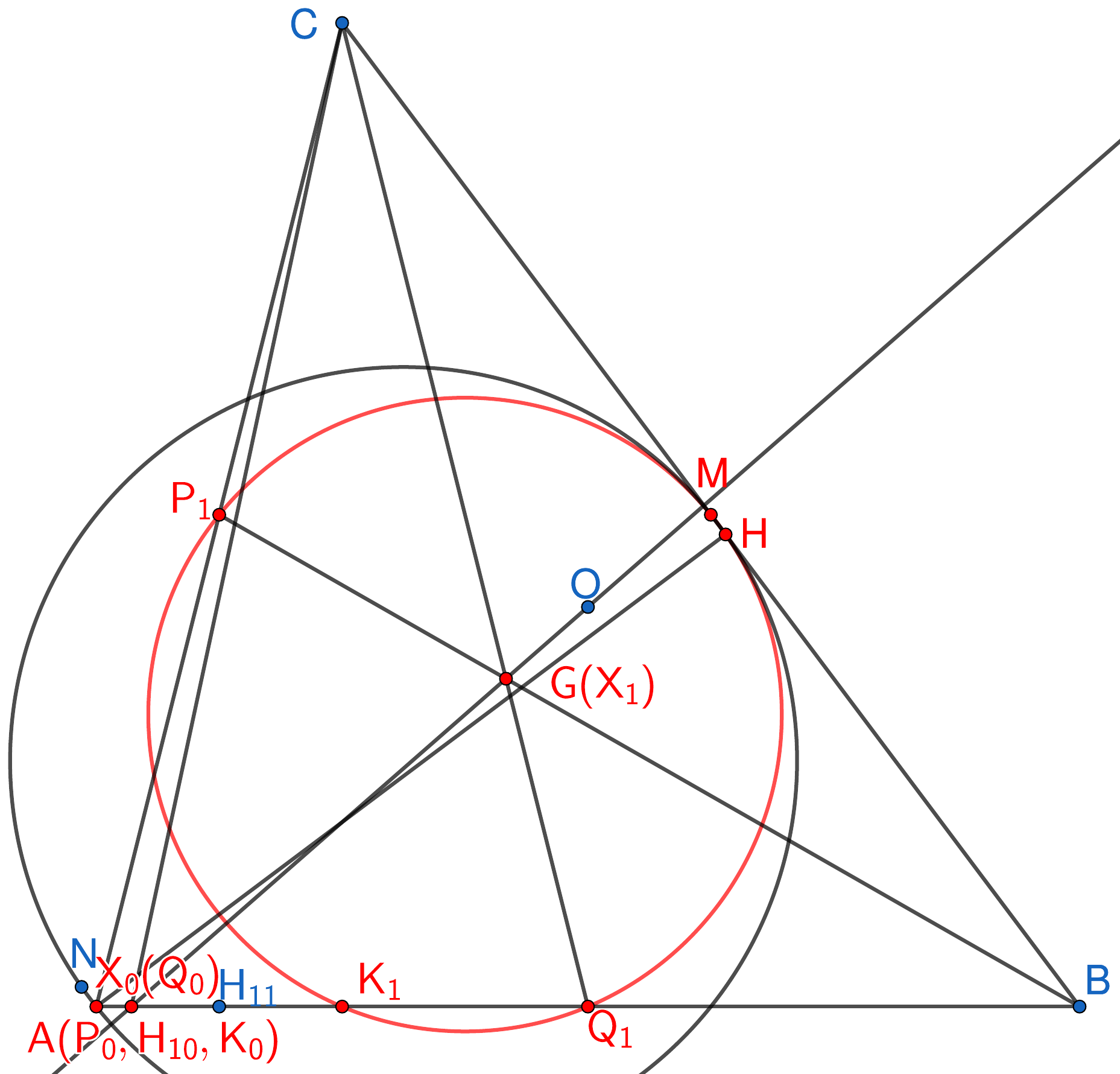}
    \caption{Counterexample to the original formalization of LEAN-IMO-Bench Advanced~010. The configuration associated with \(X_1\) (points with subscript ${}_1$) makes circumcircles of $\triangle K_1P_1Q_1$ and $\triangle P_1H_1M_1$ the same and thus forces the fixed circle to be this circumcircle, whereas the configuration associated with \(X_0\) (points with subscript ${}_0$) admits a val.id point $N$ not in that circumcircle.}
    \label{fig:010-counter-appendix}
\end{figure}

\paragraph{Counterexample and Repair} 
The intended configuration is shown in Figure~\ref{fig:010-appendix}. Figure~\ref{fig:010-counter-appendix} exhibits a
counterexample admitted by the original formalization. Consider the
configuration
\[
\begin{gathered}
A=(0,0),\qquad B=(2,0),\qquad C=(\tfrac12,2),\\
O=(1,\tfrac{13}{16}),\qquad
G=(\tfrac56,\tfrac23),\\
H=(\tfrac{32}{25},\tfrac{24}{25}),\qquad
M=(\tfrac54,1).
\end{gathered}
\]
Assume, for contradiction, that a fixed circle \(\omega\) satisfying
the original formalization exists.

First, choose \(X_1=G\). The corresponding construction gives
\[
\begin{gathered}
P_1=(\tfrac14,1),\qquad Q_1=(1,0),\\
H_{11}=(\tfrac14,0),\qquad K_1=(\tfrac12,0).
\end{gathered}
\]
Consequently, the circle through \(K_1,P_1,Q_1\) and the circle
through \(P_1,H,M\) coincide with the nine-point circle
\(\Gamma_9\) of \(\triangle ABC\). The points \(H\), \(M\), and
\(Q_1\) are all distinct from \(P_1\) and lie on both circles, so each
of them satisfies all the hypotheses imposed on \(T\) in the original
formalization. They must therefore all lie on \(\omega\). Since
\(H,M,Q_1\) are noncollinear, they determine a unique circle, and hence
\[
\omega=\Gamma_9,
\]
since \(\Gamma_9\) is the circle through \(Q_1,H,M\).

Next, choose the intersection of \(OG\) and \(AB\),
\[
X_0=(\tfrac1{14},0).
\]
For this choice, the construction degenerates:
\[
P_0=H_{10}=K_0=A,
\qquad
Q_0=X_0.
\]
Now take
\[
N=(-\tfrac3{100},\tfrac1{25}).
\]
Direct substitution verifies that \(N\) lies on the circle through
\(A,H,M\), but does not lie on \(\Gamma_9\). Moreover,
\[
\{N,K_0,P_0,Q_0\}=\{N,A,X_0\}
\]
is concyclic, since \(N,A,X_0\) are noncollinear, and
\[
\{N,P_0,H,M\}=\{N,A,H,M\}
\]
is concyclic by the choice of \(N\). Since \(N\neq P_0\), the point
\(N\) also satisfies all the hypotheses imposed on \(T\). The original
formalization would therefore imply \(N\in\omega\), contradicting
\(N\notin\Gamma_9=\omega\). Thus no such fixed circle \(\omega\)
exists.

The counterexample exploits the degenerate case that two circumcircles are the same, allowed by the
original formalization. The configuration $X_0,P_0,Q_0,H_{10},K_0$ is another kind of degenerate configuration, in which triple $KPQ$ is degenerate, while other non-degenerate configurations also contradict as long as its $T$ isn't in that circumcircle.
The phrase ``Let \(T\) be the intersection other than \(P\) of the circumcircle $\cdots$ the point \(T\)''
implicitly requires both defining triples to be noncollinear and their
circumcircles to be distinct. We therefore add these three conditions
as explicit hypotheses, as highlighted in green in the formal
statement above. With these conditions, \textsc{GeoProver} proves the
repaired statement.

\subsection{IMO 2026 P2}
\label{app:imo2026-p2}

We record the three artifacts of the IMO 2026 Problem~2 run: the informal
problem supplied to \textsc{GeoFormalizer}, the Lean~4 statement it produced,
and the proof \textsc{GeoProver} constructed for that statement.

\paragraph{Informal Statement.}
\begin{statement}
Let \(ABC\) be a triangle, and let \(M\) and \(N\) be the midpoints of
\(AB\) and \(AC\), respectively.

Let \(K\) and \(L\) be points lying strictly inside triangles \(BMC\) and
\(BNC\), respectively, such that \(K\) lies strictly inside triangle \(ABL\)
and \(L\) lies strictly inside triangle \(AKC\). Suppose that
\[
\begin{aligned}
  \angle KBA &= \angle ACL,\\
  \angle LBK &= \angle LNC,\\
  \angle LCK &= \angle BMK.
\end{aligned}
\]

Let \(O\) be the circumcenter of triangle \(AKL\). Prove that
\[
  OM=ON.
\]
\end{statement}

\paragraph{Formal Statement.}
\textsc{GeoFormalizer} deterministically translates the \textsc{GeoIR} program generated for this problem by Claude Opus~4.8 into the Mathlib-native theorem shown in Listing~\ref{lst:imo2026-statement}. Each condition that a point lies strictly inside a triangle is replaced by the weaker condition that the point belongs to the corresponding convex hull. The resulting formal theorem is therefore stronger than the original statement, and its verified proof immediately implies the original claim.


\paragraph{Formal Proof.}

Using GPT-5.6-Sol, \textsc{GeoProver} produces a 569-line Lean file in a single agent round, with a wall-clock time of 4895.4 seconds.
The resulting file compiles without \texttt{sorry}, preserves the audited statement unchanged, and depends only on  
\texttt{propext}, \texttt{Classical.choice}, and \texttt{Quot.sound}. The complete Lean source code is provided in Appendix~\ref{app:imo2026-source}.


\section{Additional Experimental Settings}
\subsection{Lean Toolchain and Computing Platform}
All generated proofs were verified on Ubuntu~24.04.3 LTS (Linux x86-64) using Lean~4.27.0 and Mathlib at commit \texttt{a3a10db}. Goedel Prover inference was served on six NVIDIA A800 GPUs using three-way data parallelism and two-way tensor parallelism (\(\mathrm{DP}=3,\ \mathrm{TP}=2\)).  Lean verification and external symbolic computation were executed on CPU and did not use the GPUs. The CPU-side workloads were executed on a machine equipped with two AMD EPYC~7H12 processors, providing 128 physical cores and 256 hardware threads, and 251\,GiB of RAM. Hilbert's retrieval encoder was served separately on an NVIDIA RTX A6000 GPU.

\subsection{Direct Formalization}
The direct formalization evaluates whether an LLM can translate an informal geometry problem directly into a Mathlib-native Lean statement. Each problem receives exactly one generation (pass@1). The original informal problem is supplied verbatim as the user message. The following system prompt is used unchanged for all seven backbones:

\begin{tcolorbox}[
  colback=gray!5,
  colframe=gray!50,
  boxrule=0.5pt
]
\small

Translate the given natural-language plane-geometry problem into a single
Lean~4 theorem. The theorem will be checked against Lean~4.27.0 with Mathlib
imported, so you may use any Mathlib definitions and notations.

\medskip

Name the theorem \texttt{problem}. End the theorem with
\texttt{:= by sorry}. Do not attempt a proof; we are evaluating the
\textsc{statement} only.

\medskip

\textbf{\# Output format}

Return a JSON object with exactly these fields:

\begin{description}
  \item[\texttt{lean}] the complete theorem as one string;
  \item[\texttt{answer}] for a ``find'' problem, the numeric or closed-form
  answer as a string; otherwise \texttt{null};
  \item[\texttt{comment}] one short sentence describing the formalization
  choices.
\end{description}

\end{tcolorbox}

\subsection{Euclean}
\label{app:baseline-euclean}

We use the authors' official release with its prompts unmodified. It formalizes
each informal problem in a single agentic run: a coding agent iterates against
the Lean compiler, reading elaborator diagnostics through a Lean language-server
tool and revising, until the statement elaborates or its ten-minute per-problem
budget is exhausted. The released repository targets Lean/Mathlib v4.28.0; we
run it on \texttt{v4.27.0} so that all systems share one toolchain, and the
geometry primitives involved are stable across these versions. Every backbone
receives the same benchmark and budget, and we decide success by re-elaborating
each generated statement in Lean ourselves rather than by the agent's
self-reported status.

We measure \emph{model calls} per problem: how many times the agent loop invokes
the LLM. Counts are taken from the agents' own
session logs, one per LLM response, over successfully formalized problems only so
that the measurement is comparable across backbones.

\begin{table}[t]
\centering
\small
\renewcommand{\arraystretch}{1.08}
\setlength{\tabcolsep}{0pt}

\begin{tabular*}{\columnwidth}{
@{\extracolsep{\fill}}
l
c
c
c
@{}
}
\toprule
Model & Mean & Median & Range \\
\midrule
GPT-5.6-Sol       & 16.8  & 17 & 7--35 \\
Claude Opus~4.8   & 14.0  & 14 & 8--26 \\
DeepSeek-V4-Pro   & 20.7  & 18 & 4--66 \\
DeepSeek-V4-Flash & 21.6  & 19 & 4--78 \\
Qwen3.7-Max       & 17.0  & 16 & 1--47 \\
MiniMax-M3        & 116.4 & 84 & 3--413 \\
GLM-5.2           & 18.4  & 14 & 3--72 \\
\bottomrule
\end{tabular*}

\caption{
Number of model calls per problem under the \textsc{Euclean} baseline
(successfully formalized problems; 10-minute budget). The 14--116 values
reported in the main text correspond to the range of mean calls across
backbones, while the maximum number of calls on a single problem is 413.
}
\label{tab:euclean-calls}
\end{table}
Every backbone averages at least 14 model calls per problem, and the loop is
unbounded within the time budget, so a single hard problem can consume hundreds
of calls.
\textsc{GeoFormalizer} instead emits one \textsc{GeoIR} program and
applies at most four bounded repair rounds, two compiler-guided and two
faithfulness-guided, for a worst case of five model calls.
In comparison, \textsc{Euclean}   requires between \(14.0\) and \(116.4\) model
calls per successfully formalized problem on average across the
evaluated backbones.

\subsection{Goedel Prover}
We evaluate the released Goedel-Prover-V2-8B checkpoint as a
sampling-based prover, which receives only the formal Lean statement. Using the released model's prompt, we draw 32 independent completions per statement (pass@32) at temperature $1.0$, top-$p$ $0.95$, and up to 16{,}384 tokens through vLLM, with no information shared across candidates. We extract the last fenced
\texttt{lean}/\texttt{lean4} block, or the whole response if none is present,
and compile it as a standalone file under
Lean~4.27.0 and the same Mathlib. A statement counts as solved only if at least one of
its 32 candidates elaborates without changing the statement and uses only the permitted Lean axioms. This evaluation of Goedel-Prover-V2-8B is separate from the official
Goedel-V2-32B leaderboard result reported in Table \ref{tab:rq3-leap}.

\subsection{Numina-Lean-Agent}
\label{app:baseline-numina}

We use the official release at commit \texttt{01d8df2} with its default
medium-mode prompt, unmodified, and its default limit of five interaction rounds
per run---an internal loop of the agent, not five independent samples. The
backbone is DeepSeek-V4-Pro; sampling parameters, including temperature and the
thinking budget, are left at the provider's defaults. Each problem receives a
two-hour wall-clock budget and runs against Lean \texttt{v4.27.0} in a sandbox
that exposes the Lean project and toolchain read-only, masks the other problems,
and disables network access.

\subsection{Hilbert}
\label{app:baseline-hilbert}

We use the asynchronous implementation at commit \texttt{bee1325} with its
released configuration, changing only the backbone: the defaults pair a 32B
fine-tuned prover with a 120B open-weight informal reasoner, and we replace both
with DeepSeek-V4-Pro so that the comparison isolates the proving framework
rather than the underlying model. Prompts, retrieval, proof checking, sampling
parameters, and the agent's internal recursion and repair limits are all left as
released; the implementation fixes temperature at $0.6$. Each problem receives
one pass@1 run against Lean \texttt{v4.27.0} under the same two-hour
budget, within which its model calls are not otherwise capped. This modified Hilbert configuration is separate from the official Hilbert
leaderboard result reported in Table \ref{tab:rq3-leap}.
\clearpage
\begin{leancode}[\lstset{label=lst:imo2026-statement, caption={The Lean~4 statement of IMO 2026 P2 produced by
\textsc{GeoFormalizer}, reproduced verbatim from the checked file.  The proof
term supplied by \textsc{GeoProver} closes this goal without altering any
hypotheses.}}]
theorem Midpoints_Interior_AKL :
    ∀ A : EuclideanSpace ℝ (Fin 2),
    ∀ B : EuclideanSpace ℝ (Fin 2),
    ∀ C : EuclideanSpace ℝ (Fin 2),
    AffineIndependent (P := EuclideanSpace ℝ (Fin 2)) ℝ ![A, B, C] →
    let M := midpoint (P := EuclideanSpace ℝ (Fin 2)) ℝ A B;
    let N := midpoint (P := EuclideanSpace ℝ (Fin 2)) ℝ A C;
    ∀ K : EuclideanSpace ℝ (Fin 2),
    ∀ L : EuclideanSpace ℝ (Fin 2),
    AffineIndependent (P := EuclideanSpace ℝ (Fin 2)) ℝ ![B, M, C] →
    AffineIndependent (P := EuclideanSpace ℝ (Fin 2)) ℝ ![B, N, C] →
    K ∈ convexHull ℝ ({B, M, C} : Set (EuclideanSpace ℝ (Fin 2))) →
    L ∈ convexHull ℝ ({B, N, C} : Set (EuclideanSpace ℝ (Fin 2))) →
    AffineIndependent (P := EuclideanSpace ℝ (Fin 2)) ℝ ![A, B, L] →
    AffineIndependent (P := EuclideanSpace ℝ (Fin 2)) ℝ ![A, K, C] →
    K ∈ convexHull ℝ ({A, B, L} : Set (EuclideanSpace ℝ (Fin 2))) →
    L ∈ convexHull ℝ ({A, K, C} : Set (EuclideanSpace ℝ (Fin 2))) →
    (K ≠ B ∧ B ≠ A) →
    (A ≠ C ∧ C ≠ L) →
    (L ≠ B ∧ B ≠ K) →
    (L ≠ N ∧ N ≠ C) →
    (L ≠ C ∧ C ≠ K) →
    (B ≠ M ∧ M ≠ K) →
    (EuclideanGeometry.angle (P := EuclideanSpace ℝ (Fin 2)) K B A) =
      (EuclideanGeometry.angle (P := EuclideanSpace ℝ (Fin 2)) A C L) →
    (EuclideanGeometry.angle (P := EuclideanSpace ℝ (Fin 2)) L B K) =
      (EuclideanGeometry.angle (P := EuclideanSpace ℝ (Fin 2)) L N C) →
    (EuclideanGeometry.angle (P := EuclideanSpace ℝ (Fin 2)) L C K) =
      (EuclideanGeometry.angle (P := EuclideanSpace ℝ (Fin 2)) B M K) →
    AffineIndependent (P := EuclideanSpace ℝ (Fin 2)) ℝ ![A, K, L] →
    ∀ O : EuclideanSpace ℝ (Fin 2),
    (Dist.dist (α := EuclideanSpace ℝ (Fin 2)) O A) =
      (Dist.dist (α := EuclideanSpace ℝ (Fin 2)) O K) →
    (Dist.dist (α := EuclideanSpace ℝ (Fin 2)) O A) =
      (Dist.dist (α := EuclideanSpace ℝ (Fin 2)) O L) →
    ((Dist.dist (α := EuclideanSpace ℝ (Fin 2)) O M) =
      (Dist.dist (α := EuclideanSpace ℝ (Fin 2)) O N))
\end{leancode}

\clearpage

\newcommand{\rowsep}{\\ \arrayrulecolor{lightgray}\midrule[0.1pt]\arrayrulecolor{black}}

\begingroup
\footnotesize
\setlength{\tabcolsep}{4pt}
\renewcommand{\arraystretch}{1.15}
\begin{longtable}{@{}>{\raggedright\arraybackslash}p{0.32\textwidth}>{\raggedright\arraybackslash}p{0.45\textwidth}>{\raggedright\arraybackslash}p{0.23\textwidth}@{}}
\caption{GeoIR--Lean Specification}\label{tab:geodsl-lean}\\
\multicolumn{3}{l}{The IR notations suffixed with \lean{!} are translated into Lean with non-degenerate conditions}\\
\toprule
\textbf{IR} & \textbf{Lean} & \textbf{Semantics}\\
\midrule
\endfirsthead
\multicolumn{3}{@{}l}{\footnotesize\emph{Table~\ref{tab:geodsl-lean} (continued)}}\\[2pt]
\toprule
\textbf{IR} & \textbf{Lean} & \textbf{Semantics}\\
\midrule
\endhead
\midrule
\multicolumn{3}{r@{}}{\footnotesize\emph{continued on next page}}\\
\endfoot
\bottomrule
\endlastfoot
\multicolumn{3}{@{}l}{\textbf{0.~Statements and program structure}}\\*
\midrule
\lstinline"point A ;" & \lean{(A : EuclideanSpace ℝ (Fin 2))} & introduce a free point of the plane (a binder of the theorem) \rowsep
\lstinline"real r ;" & \lean{(r : ℝ)} & introduce a free real parameter (a binder) \rowsep
\lstinline"circle S ;" & \lean{(S : EuclideanGeometry.Sphere (EuclideanSpace ℝ (Fin 2)))} & introduce a free circle (a binder) \rowsep
\lstinline"let M = <expr> ;" & \lean{let M := <expr>} & denote a point, a number, or a circle as $M$ \rowsep
\lstinline"incircle S of A B C ; ..." & \lean{∀ h : AffineIndependent (P := EuclideanSpace ℝ (Fin 2)) ℝ ![A, B, C], let S := Affine.Simplex.insphere (P := EuclideanSpace ℝ (Fin 2)) ⟨![A, B, C], h⟩; ...} & name the incircle of triangle $ABC$ \rowsep
\lstinline"assume <prop> ;" & \lean{<prop> →} & a hypothesis; an implication premise before the conclusion \rowsep
\lstinline"prove <prop>" & \lean{prop} (last statement) \\
\midrule
\multicolumn{3}{@{}l}{\textbf{1.~Point constructors (values in the type of points)}}\\*
\midrule
\lstinline"( x , y )" & \lean{!₂[(x : ℝ), y]} & the point with explicit coordinates \rowsep
\lstinline"midpoint A B" & \lean{midpoint (P := EuclideanSpace ℝ (Fin 2)) ℝ A B} & midpoint of segment $AB$ \rowsep
\lstinline"centroid A B C" & \lean{Finset.centroid (P := EuclideanSpace ℝ (Fin 2)) ℝ (Finset.univ (α := Fin 3)) ![A, B, C]} & centroid of triangle $ABC$ \rowsep
\lstinline"lerp t A B" & \lean{AffineMap.lineMap (P1 := EuclideanSpace ℝ (Fin 2)) A B (t : ℝ)} & the point $A + t\,(B - A)$ on line $AB$, $t \in \mathbb{R}$ \rowsep
\lstinline"divide A B m n" & \lean{AffineMap.lineMap (k := ℝ) (P1 := EuclideanSpace ℝ (Fin 2)) A B (m / (m + n) : ℝ)} & the point $P$ on $AB$ with $\mathit{AP} : \mathit{PB} = m : n$ \rowsep
\lstinline"center S" & \lean{EuclideanGeometry.Sphere.center (S : EuclideanGeometry.Sphere (EuclideanSpace ℝ (Fin 2)))} & the centre of circle $S$ \\
\midrule
\multicolumn{3}{@{}l}{\textbf{2.~Circles (values in the type of circles)}}\\*
\midrule
\lstinline"circle O r" & \lean{(⟨O, r⟩ : EuclideanGeometry.Sphere (EuclideanSpace ℝ (Fin 2)))} & the circle with centre $O$ and radius $r$ \\
\midrule
\multicolumn{3}{@{}l}{\textbf{3.~Predicates (propositions)}}\\*
\multicolumn{3}{@{}l}{\emph{3.1~About points and lines (arguments are points)}}\\*
\midrule
\lstinline"collinear A B C" & \lean"Collinear ℝ {A, B, C}" & $A, B, C$ lie on one line \rowsep
\lstinline"noncollinear A B C" & \lean{AffineIndependent (P := EuclideanSpace ℝ (Fin 2)) ℝ ![A, B, C]} & $A, B, C$ form a genuine triangle (not collinear) \rowsep
\lstinline"parallel A B C D" & \lean{¬ LinearIndependent (M := EuclideanSpace ℝ (Fin 2)) ℝ ![A -ᵥ B, C -ᵥ D]} & line $AB$ is parallel to line $CD$ (degenerate lines allowed) \rowsep
\lstinline"nondegenerate_of_parallel A B C D" & \lean{(A ≠ B ∧ C ≠ D)} & non-degeneracy of \lstinline"parallel A B C D" \rowsep
\lstinline"parallel! A B C D" & \lean{(A ≠ B ∧ C ≠ D ∧ ¬ LinearIndependent (M := EuclideanSpace ℝ (Fin 2)) ℝ ![A -ᵥ B, C -ᵥ D])} & parallel with the non-degeneracy conditions built in \rowsep
\lstinline"perp A B C D" & \lean{Inner.inner (E := EuclideanSpace ℝ (Fin 2)) ℝ (A -ᵥ B) (C -ᵥ D) = 0} & line $AB$ is perpendicular to line $CD$ \rowsep
\lstinline"perp! A B C D" & \lean{(A ≠ B ∧ C ≠ D ∧ Inner.inner (E := EuclideanSpace ℝ (Fin 2)) ℝ (A -ᵥ B) (C -ᵥ D) = 0)} & perpendicular with the non-degeneracy conditions built in \rowsep
\lstinline"right_angle A B C" & \lean{Inner.inner (E := EuclideanSpace ℝ (Fin 2)) ℝ (A -ᵥ B) (C -ᵥ B) = 0} & the angle at $B$ (in $ABC$) is $90^\circ$ \rowsep
\lstinline"nondegenerate_of_right_angle A B C" & \lean{(A ≠ B ∧ C ≠ B)} & non-degeneracy of a right angle $ABC$ \rowsep
\lstinline"right_angle! A B C" & \lean{(A ≠ B ∧ C ≠ B ∧ Inner.inner (E := EuclideanSpace ℝ (Fin 2)) ℝ (A -ᵥ B) (C -ᵥ B) = 0)} & right angle with the non-degeneracy conditions built in \rowsep
\lstinline"parallelogram A B C D" & \lean{(¬ LinearIndependent (M := EuclideanSpace ℝ (Fin 2)) ℝ ![A -ᵥ B, C -ᵥ D] ∧ ¬ LinearIndependent (M := EuclideanSpace ℝ (Fin 2)) ℝ ![A -ᵥ D, C -ᵥ B])} & $ABCD$ is a parallelogram ($AB \parallel CD$ and $AD \parallel CB$) \rowsep
\lstinline"nondegenerate_of_parallelogram A B C D" & \lean{AffineIndependent (P := EuclideanSpace ℝ (Fin 2)) ℝ ![A, B, C]} & non-degeneracy of a parallelogram $ABCD$ \rowsep
\lstinline"parallelogram! A B C D" & \lean{(AffineIndependent (P := EuclideanSpace ℝ (Fin 2)) ℝ ![A, B, C] ∧ ¬ LinearIndependent (M := EuclideanSpace ℝ (Fin 2)) ℝ ![A -ᵥ B, C -ᵥ D] ∧ ¬ LinearIndependent (M := EuclideanSpace ℝ (Fin 2)) ℝ ![A -ᵥ D, C -ᵥ B])} & parallelogram with the non-degeneracy conditions built in \rowsep
\lstinline"concyclic A B C D" & \lean"EuclideanGeometry.Concyclic (P := EuclideanSpace ℝ (Fin 2)) {A, B, C, D}" & $A, B, C, D$ lie on one circle (no auxiliary centre needed) \rowsep
\lstinline"are P A B C concyclic_with_center O" & \lean{(dist P O = dist A O ∧ dist P O = dist B O ∧ dist P O = dist C O)} & $P, A, B, C$ on a circle centred at $O$ \rowsep
\lstinline"is X weakly_between A B" & \lean{Wbtw ℝ A X B} & $X$ on segment $AB$, endpoints allowed \rowsep
\lstinline"is X strictly_between A B" & \lean{Sbtw ℝ A X B} & $X$ strictly inside segment $AB$ \rowsep
\lstinline"is X in_convex_hull P1 ... Pn" & \lean"X ∈ convexHull ℝ ({P1, ..., Pn} : Set (EuclideanSpace ℝ (Fin 2)))" & $X$ inside the convex hull of $P_1,\dots,P_n$ ($n \le 6$) \rowsep
\lstinline"nondegenerate_of_line A B" & \lean{A ≠ B} & non-degeneracy of line/segment $AB$: $A \neq B$ \rowsep
\lstinline"nondegenerate_of_triangle A B C" & \lean{AffineIndependent (P := EuclideanSpace ℝ (Fin 2)) ℝ ![A, B, C]} & non-degeneracy of triangle $ABC$ (same as \lstinline"noncollinear") \rowsep
\lstinline"nondegenerate_of_dist_line A B" & \lean{A ≠ B} & non-degeneracy of \lstinline"dist_line P A B": $A \neq B$ \rowsep
\lstinline"nondegenerate_of_angle A B C" & \lean{(A ≠ B ∧ B ≠ C)} & non-degeneracy of angle $ABC$ \rowsep
\lstinline"convex_quad A B C D" & \lean{IsConvexQuad A B C D} & $ABCD$ is a convex quadrilateral (vertices in cyclic order) \rowsep
\lstinline"nondegenerate_of_incircle A B C" & \lean{AffineIndependent (P := EuclideanSpace ℝ (Fin 2)) ℝ ![A, B, C]} & non-degeneracy of the incircle of $ABC$: a genuine triangle \\
\midrule
\multicolumn{3}{@{}l}{\emph{3.2~About circles}}\\*
\midrule
\lstinline"nondegenerate_circle r" & \lean{(r : ℝ) > 0} & non-degeneracy of a circle of radius $r$: $r > 0$ \rowsep
\lstinline"is_tangent S A B" & \lean"EuclideanGeometry.Sphere.IsTangent (S : EuclideanGeometry.Sphere (EuclideanSpace ℝ (Fin 2))) (affineSpan ℝ {A, B})" & line $AB$ is tangent to circle $S$ \\
\midrule
\multicolumn{3}{@{}l}{\emph{3.3~Comparisons (points, circles, or real numbers)}}\\*
\midrule
\lstinline"(a = b)" & \lean{(a = b)} & equality \rowsep
\lstinline"(a != b)" & \lean{(a ≠ b)} & disequality \\
\midrule
\multicolumn{3}{@{}l}{\emph{3.4~Comparisons (real numbers)}}\\*
\midrule
\lstinline"(a <= b)" & \lean{(a ≤ b)} & less than or equal \rowsep
\lstinline"(a > b)" & \lean{(a > b)} & greater than \rowsep
\lstinline"(a >= b)" & \lean{(a ≥ b)} & greater than or equal \rowsep
\lstinline"(a < b)" & \lean{(a < b)} & less than \\
\midrule
\multicolumn{3}{@{}l}{\emph{3.5~Logic connectives}}\\*
\midrule
\lstinline"(p and q)" & \lean{(p ∧ q)} & conjunction \rowsep
\lstinline"(p or q)" & \lean{(p ∨ q)} & disjunction \rowsep
\lstinline"(not p)" & \lean{¬ p} & negation \rowsep
\lstinline"(p iff q)" & \lean{(p ↔ q)} & biconditional \rowsep
\lstinline"(p -> q)" & \lean{(p → q)} & implication \\
\midrule
\multicolumn{3}{@{}l}{\textbf{4.~Functions (values in the type of real numbers)}}\\*
\multicolumn{3}{@{}l}{\emph{4.1~Arguments are points}}\\*
\midrule
\lstinline"dist A B" & \lean{Dist.dist (α := EuclideanSpace ℝ (Fin 2)) A B} & distance $\lvert AB\rvert$; squared form \lstinline"sq (dist A B)" vs.\ \lean{dist A B ^ 2} \rowsep
\lstinline"dot A B C D" & \lean{Inner.inner (E := EuclideanSpace ℝ (Fin 2)) ℝ (A -ᵥ B) (C -ᵥ D)} & dot product of vectors $AB$ and $CD$ \rowsep
\lstinline"dist_line P A B" & \lean"Dist.dist P ((EuclideanGeometry.orthogonalProjection (V := EuclideanSpace ℝ (Fin 2)) (affineSpan ℝ {A, B})) P)" & distance from point $P$ to line $AB$ \rowsep
\lstinline"triangle_area A B C" & \lean{(|(Module.Basis.orientation <| PiLp.basisFun 2 ℝ (Fin 2)).areaForm (A -ᵥ B) (C -ᵥ B)| / 2)} & the (unsigned) area of triangle $ABC$ \rowsep
\lstinline"xof P" & \lean{(P : EuclideanSpace ℝ (Fin 2)).ofLp 0} & $x$-coordinate of a point $P$ \rowsep
\lstinline"yof P" & \lean{(P : EuclideanSpace ℝ (Fin 2)).ofLp 1} & $y$-coordinate of a point $P$ \rowsep
\lstinline"angle A B C" & \lean{EuclideanGeometry.angle (P := EuclideanSpace ℝ (Fin 2)) A B C} & the undirected angle at $B$, in radians, in $[0, \pi]$ \\
\midrule
\multicolumn{3}{@{}l}{\emph{4.2~Real number constants}}\\*
\midrule
\lstinline"pi" & \lean{Real.pi} & the constant $\pi$ (angles in radians) \rowsep
natural number literal, e.g.\ \lstinline"2" & \lean{(2 : ℝ)} & any natural literal; numerals are real numbers \\
\midrule
\multicolumn{3}{@{}l}{\emph{4.3~Angle and trigonometry}}\\*
\midrule
\lstinline"degree d" & \lean{(Real.pi * (d : ℝ) / 180)} & $d$ degrees, converted to radians (e.g.\ \lstinline"degree 90" $= \pi / 2$) \rowsep
\lstinline"sin a" & \lean{Real.sin a} & sine (of a radian value, e.g.\ \lstinline"sin (angle A B C)") \rowsep
\lstinline"cos a" & \lean{Real.cos a} & cosine (of a radian value, e.g.\ \lstinline"cos (angle A B C)") \rowsep
\lstinline"sqrt a" & \lean{Real.sqrt a} & real square root \rowsep
\lstinline"sq a" & \lean{(a : ℝ) ^ 2} & the square of $a$ \rowsep
\lstinline"abs a" & \lean{|(a : ℝ)|} & absolute value \rowsep
\lstinline"radius S" & \lean{EuclideanGeometry.Sphere.radius (S : EuclideanGeometry.Sphere (EuclideanSpace ℝ (Fin 2)))} & the radius of circle $S$ \\
\midrule
\multicolumn{3}{@{}l}{\emph{4.4~Arithmetic}}\\*
\midrule
\lstinline"a + b" & \lean{a + b} & addition \rowsep
\lstinline"a - b" & \lean{a - b} & subtraction \rowsep
\lstinline"a * b" & \lean{a * b} & multiplication \rowsep
\lstinline"a / b" & \lean{a / b} & division \rowsep
\lstinline"(- a)" & \lean{-a} & unary minus \\
\lstinline"a ^ 2" & \lean{a ^ 2} & $a ^ 2$ \\
\midrule
\multicolumn{3}{@{}l}{\textbf{5.~Others}}\\*
\midrule
\lstinline"( e )" & \lean{( e )} & explicit priority; redundant parentheses are always allowed \\
\end{longtable}
\endgroup

\clearpage



\section{A Complete Lean Proof of IMO 2026 P2}
\label{app:imo2026-source}

The problem statement is provided in Appendix~\ref{app:imo2026-p2}.


\newcommand{\bbk}{\ensuremath{\Bbbk}}
\begin{leancode}[\lstset{basicstyle=\scriptsize\ttfamily,numbers=left,numberstyle=\tiny,numbersep=4pt,xleftmargin=1.7em}]
import Mathlib
import GeoTheorem.ToPolyMacros

set_option maxHeartbeats 5000000
set_option maxRecDepth 100000

private def dt (a b c d : ℝ) : ℝ := a*c+b*d
private def cr (a b c d : ℝ) : ℝ := a*d-b*c
private def ori (A B C : EuclideanSpace ℝ (Fin 2)) : ℝ :=
  cr (B 0-A 0) (B 1-A 1) (C 0-A 0) (C 1-A 1)

private theorem side_nonneg_of_mem_triangle
    (A B C X : EuclideanSpace ℝ (Fin 2))
    (hX : X ∈ convexHull ℝ ({A,B,C} : Set (EuclideanSpace ℝ (Fin 2)))) :
    0 ≤ ori A B C * ori A B X := by
  apply (convexHull_min ((*\bbk*) := ℝ) ?_ ?_) hX
  · intro Z hZ
    simp only [Set.mem_insert_iff, Set.mem_singleton_iff] at hZ
    rcases hZ with hZ | hZ | hZ
    · subst Z
      change 0 ≤ ori A B C * ori A B A
      simp [ori, cr]
    · subst Z
      change 0 ≤ ori A B C * ori A B B
      dsimp [ori, cr]
      ring_nf
      norm_num
    · subst Z
      change 0 ≤ ori A B C * ori A B C
      simpa [pow_two] using sq_nonneg (ori A B C)
  · rw [convex_iff_add_mem]
    intro P hP Q hQ a b ha hb hab
    change 0 ≤ ori A B C * ori A B P at hP
    change 0 ≤ ori A B C * ori A B Q at hQ
    change 0 ≤ ori A B C * ori A B (a • P + b • Q)
    have hor :
        ori A B (a • P + b • Q) = a * ori A B P + b * ori A B Q := by
      have hb' : b = 1-a := by linarith
      rw [hb']
      dsimp [ori, cr]
      ring
    rw [hor, mul_add]
    simpa only [mul_assoc, mul_comm, mul_left_comm] using
      add_nonneg (mul_nonneg ha hP) (mul_nonneg hb hQ)

private theorem ori_ne_zero_of_affineIndependent
    (A B C : EuclideanSpace ℝ (Fin 2))
    (h : AffineIndependent ℝ ![A,B,C]) : ori A B C ≠ 0 := by
  to_poly at h
  dsimp [ori, cr]
  intro hz
  apply h
  linear_combination -hz

private theorem same_sign_of_oriented
    (D a b : ℝ) (hD : D ≠ 0) (ha : 0 ≤ D*a) (hb : 0 ≤ D*b) :
    (0 ≤ a ∧ 0 ≤ b) ∨ (a ≤ 0 ∧ b ≤ 0) := by
  rcases lt_or_gt_of_ne hD with hD | hD
  · right
    constructor <;> nlinarith
  · left
    constructor <;> nlinarith

private theorem oriented_trans
    (D E a : ℝ) (hE : E ≠ 0) (hDE : 0 ≤ D*E) (hEa : 0 ≤ E*a) :
    0 ≤ D*a := by
  rcases lt_or_gt_of_ne hE with hE | hE
  · have hD : D ≤ 0 := by nlinarith
    have ha : a ≤ 0 := by nlinarith
    exact mul_nonneg_of_nonpos_of_nonpos hD ha
  · have hD : 0 ≤ D := by nlinarith
    have ha : 0 ≤ a := by nlinarith
    exact mul_nonneg hD ha

private theorem three_angle_cross_signs
    (A B C K L : EuclideanSpace ℝ (Fin 2))
    (hABC : AffineIndependent ℝ ![A,B,C])
    (hABL : AffineIndependent ℝ ![A,B,L])
    (hAKC : AffineIndependent ℝ ![A,K,C])
    (hK₁ : K ∈ convexHull ℝ ({B, midpoint ℝ A B, C} :
      Set (EuclideanSpace ℝ (Fin 2))))
    (hL₁ : L ∈ convexHull ℝ ({B, midpoint ℝ A C, C} :
      Set (EuclideanSpace ℝ (Fin 2))))
    (hK₂ : K ∈ convexHull ℝ ({A,B,L} : Set (EuclideanSpace ℝ (Fin 2))))
    (hL₂ : L ∈ convexHull ℝ ({A,K,C} : Set (EuclideanSpace ℝ (Fin 2)))) :
    let M := midpoint ℝ A B
    let N := midpoint ℝ A C
    ((0 ≤ ori B K A ∧ 0 ≤ ori C A L) ∨
      (ori B K A ≤ 0 ∧ ori C A L ≤ 0)) ∧
    ((0 ≤ ori B L K ∧ 0 ≤ ori N L C) ∨
      (ori B L K ≤ 0 ∧ ori N L C ≤ 0)) ∧
    ((0 ≤ ori C L K ∧ 0 ≤ ori M B K) ∨
      (ori C L K ≤ 0 ∧ ori M B K ≤ 0)) := by
  dsimp only
  let M := midpoint ℝ A B
  let N := midpoint ℝ A C
  let D := ori A B C
  have hD : D ≠ 0 := ori_ne_zero_of_affineIndependent A B C hABC
  have hAabc : A ∈ convexHull ℝ ({A,B,C} : Set (EuclideanSpace ℝ (Fin 2))) :=
    subset_convexHull ℝ _ (by simp)
  have hBabc : B ∈ convexHull ℝ ({A,B,C} : Set (EuclideanSpace ℝ (Fin 2))) :=
    subset_convexHull ℝ _ (by simp)
  have hCabc : C ∈ convexHull ℝ ({A,B,C} : Set (EuclideanSpace ℝ (Fin 2))) :=
    subset_convexHull ℝ _ (by simp)
  have hMabc : M ∈ convexHull ℝ ({A,B,C} : Set (EuclideanSpace ℝ (Fin 2))) := by
    exact (convex_convexHull ℝ _).midpoint_mem hAabc hBabc
  have hNabc : N ∈ convexHull ℝ ({A,B,C} : Set (EuclideanSpace ℝ (Fin 2))) := by
    exact (convex_convexHull ℝ _).midpoint_mem hAabc hCabc
  have hKabc : K ∈ convexHull ℝ ({A,B,C} : Set (EuclideanSpace ℝ (Fin 2))) := by
    apply (convexHull_min ((*\bbk*) := ℝ) ?_ (convex_convexHull ℝ _)) hK₁
    intro Z hZ
    simp only [Set.mem_insert_iff, Set.mem_singleton_iff] at hZ
    rcases hZ with hZ | hZ | hZ
    · simpa [hZ] using hBabc
    · simpa [M, hZ] using hMabc
    · simpa [hZ] using hCabc
  have hLabc : L ∈ convexHull ℝ ({A,B,C} : Set (EuclideanSpace ℝ (Fin 2))) := by
    apply (convexHull_min ((*\bbk*) := ℝ) ?_ (convex_convexHull ℝ _)) hL₁
    intro Z hZ
    simp only [Set.mem_insert_iff, Set.mem_singleton_iff] at hZ
    rcases hZ with hZ | hZ | hZ
    · simpa [hZ] using hBabc
    · simpa [N, hZ] using hNabc
    · simpa [hZ] using hCabc
  have h11 : 0 ≤ D * ori B K A := by
    have hh := side_nonneg_of_mem_triangle B M C K (by simpa [M] using hK₁)
    dsimp [D, M, ori, cr] at hh ⊢
    to_poly at hh
    nlinarith
  have h12 : 0 ≤ D * ori C A L := by
    have hperm : L ∈ convexHull ℝ ({C,N,B} : Set (EuclideanSpace ℝ (Fin 2))) := by
      rw [show ({C,N,B} : Set (EuclideanSpace ℝ (Fin 2))) =
          {B, midpoint ℝ A C, C} by
        ext Z
        simp only [N, Set.mem_insert_iff, Set.mem_singleton_iff]
        aesop]
      exact hL₁
    have hh := side_nonneg_of_mem_triangle C N B L hperm
    dsimp [D, N, ori, cr] at hh ⊢
    to_poly at hh
    nlinarith
  have h21 : 0 ≤ D * ori B L K := by
    have hDE := side_nonneg_of_mem_triangle A B C L hLabc
    have hperm : K ∈ convexHull ℝ ({B,L,A} : Set (EuclideanSpace ℝ (Fin 2))) := by
      rw [show ({B,L,A} : Set (EuclideanSpace ℝ (Fin 2))) = {A,B,L} by
        ext Z
        simp only [Set.mem_insert_iff, Set.mem_singleton_iff]
        aesop]
      exact hK₂
    have hEa := side_nonneg_of_mem_triangle B L A K hperm
    have hE : ori A B L ≠ 0 := ori_ne_zero_of_affineIndependent A B L hABL
    apply oriented_trans D (ori A B L) (ori B L K) hE
    · simpa [D] using hDE
    · dsimp [ori, cr] at hEa ⊢
      nlinarith
  have h22 : 0 ≤ D * ori N L C := by
    have hperm : L ∈ convexHull ℝ ({N,C,B} : Set (EuclideanSpace ℝ (Fin 2))) := by
      rw [show ({N,C,B} : Set (EuclideanSpace ℝ (Fin 2))) =
          {B, midpoint ℝ A C, C} by
        ext Z
        simp only [N, Set.mem_insert_iff, Set.mem_singleton_iff]
        aesop]
      exact hL₁
    have hh := side_nonneg_of_mem_triangle N C B L hperm
    dsimp [D, N, ori, cr] at hh ⊢
    to_poly at hh ⊢
    nlinarith
  have h31 : 0 ≤ D * ori C L K := by
    have hpermK : K ∈ convexHull ℝ ({A,C,B} : Set (EuclideanSpace ℝ (Fin 2))) := by
      rw [show ({A,C,B} : Set (EuclideanSpace ℝ (Fin 2))) = {A,B,C} by
        ext Z
        simp only [Set.mem_insert_iff, Set.mem_singleton_iff]
        aesop]
      exact hKabc
    have hDE₀ := side_nonneg_of_mem_triangle A C B K hpermK
    have hpermL : L ∈ convexHull ℝ ({C,K,A} : Set (EuclideanSpace ℝ (Fin 2))) := by
      rw [show ({C,K,A} : Set (EuclideanSpace ℝ (Fin 2))) = {A,K,C} by
        ext Z
        simp only [Set.mem_insert_iff, Set.mem_singleton_iff]
        aesop]
      exact hL₂
    have hEa₀ := side_nonneg_of_mem_triangle C K A L hpermL
    have hE : ori A K C ≠ 0 := ori_ne_zero_of_affineIndependent A K C hAKC
    apply oriented_trans D (ori A K C) (ori C L K) hE
    · dsimp [D, ori, cr] at hDE₀ ⊢
      nlinarith
    · dsimp [ori, cr] at hEa₀ ⊢
      nlinarith
  have h32 : 0 ≤ D * ori M B K := by
    have hperm : K ∈ convexHull ℝ ({M,B,C} : Set (EuclideanSpace ℝ (Fin 2))) := by
      rw [show ({M,B,C} : Set (EuclideanSpace ℝ (Fin 2))) =
          {B, midpoint ℝ A B, C} by
        ext Z
        simp only [M, Set.mem_insert_iff, Set.mem_singleton_iff]
        aesop]
      exact hK₁
    have hh := side_nonneg_of_mem_triangle M B C K hperm
    dsimp [D, M, ori, cr] at hh ⊢
    to_poly at hh ⊢
    nlinarith
  exact ⟨same_sign_of_oriented D _ _ hD h11 h12,
    same_sign_of_oriented D _ _ hD h21 h22,
    same_sign_of_oriented D _ _ hD h31 h32⟩

private theorem angle_tangent_of_same_sign
    (P Q R S T U : EuclideanSpace ℝ (Fin 2))
    (hPQ : P ≠ Q) (hRQ : R ≠ Q) (hST : S ≠ T) (hUT : U ≠ T)
    (ha : EuclideanGeometry.angle P Q R = EuclideanGeometry.angle S T U)
    (hsign :
      (0 ≤ cr (P 0-Q 0) (P 1-Q 1) (R 0-Q 0) (R 1-Q 1) ∧
       0 ≤ cr (S 0-T 0) (S 1-T 1) (U 0-T 0) (U 1-T 1)) ∨
      (cr (P 0-Q 0) (P 1-Q 1) (R 0-Q 0) (R 1-Q 1) ≤ 0 ∧
       cr (S 0-T 0) (S 1-T 1) (U 0-T 0) (U 1-T 1) ≤ 0)) :
    cr (P 0-Q 0) (P 1-Q 1) (R 0-Q 0) (R 1-Q 1) *
        dt (S 0-T 0) (S 1-T 1) (U 0-T 0) (U 1-T 1) -
      dt (P 0-Q 0) (P 1-Q 1) (R 0-Q 0) (R 1-Q 1) *
        cr (S 0-T 0) (S 1-T 1) (U 0-T 0) (U 1-T 1) = 0 := by
  have hc := congrArg Real.cos ha
  have hs := congrArg Real.sin ha
  simp only [ToPoly.cos_angle', ToPoly.inner, ToPoly.vsub_apply] at hc
  rw [ToPoly.sin_angle' hPQ hRQ, ToPoly.sin_angle' hST hUT] at hs
  simp only [ToPoly.vsub_apply] at hs
  have hnPQ : ‖P -ᵥ Q‖ ≠ 0 := norm_ne_zero_iff.mpr (vsub_ne_zero.mpr hPQ)
  have hnRQ : ‖R -ᵥ Q‖ ≠ 0 := norm_ne_zero_iff.mpr (vsub_ne_zero.mpr hRQ)
  have hnST : ‖S -ᵥ T‖ ≠ 0 := norm_ne_zero_iff.mpr (vsub_ne_zero.mpr hST)
  have hnUT : ‖U -ᵥ T‖ ≠ 0 := norm_ne_zero_iff.mpr (vsub_ne_zero.mpr hUT)
  field_simp [hnPQ, hnRQ, hnST, hnUT] at hc hs
  have hs' :
      ‖S -ᵥ T‖ * ‖U -ᵥ T‖ *
          |cr (P 0-Q 0) (P 1-Q 1) (R 0-Q 0) (R 1-Q 1)| =
        ‖P -ᵥ Q‖ * ‖R -ᵥ Q‖ *
          |cr (S 0-T 0) (S 1-T 1) (U 0-T 0) (U 1-T 1)| := by
    simpa only [cr, mul_comm, mul_left_comm, mul_assoc] using hs
  have hs'' :
      ‖S -ᵥ T‖ * ‖U -ᵥ T‖ *
          cr (P 0-Q 0) (P 1-Q 1) (R 0-Q 0) (R 1-Q 1) =
        ‖P -ᵥ Q‖ * ‖R -ᵥ Q‖ *
          cr (S 0-T 0) (S 1-T 1) (U 0-T 0) (U 1-T 1) := by
    rcases hsign with hsign | hsign
    · simpa [abs_of_nonneg hsign.1, abs_of_nonneg hsign.2] using hs'
    · rw [abs_of_nonpos hsign.1, abs_of_nonpos hsign.2] at hs'
      linear_combination -hs'
  dsimp [dt, cr] at *
  have hn : ‖S -ᵥ T‖ * ‖U -ᵥ T‖ ≠ 0 := mul_ne_zero hnST hnUT
  apply (mul_eq_zero.mp ?_).resolve_left hn
  linear_combination
    ( ((S 0 - T 0) * (U 0 - T 0) + (S 1 - T 1) * (U 1 - T 1)) ) * hs'' -
    ( ((S 0 - T 0) * (U 1 - T 1) - (S 1 - T 1) * (U 0 - T 0)) ) * hc

private def cx (A B X : EuclideanSpace ℝ (Fin 2)) : ℝ :=
  dt (B 0-A 0) (B 1-A 1) (X 0-A 0) (X 1-A 1)

private def cy (A B X : EuclideanSpace ℝ (Fin 2)) : ℝ :=
  cr (B 0-A 0) (B 1-A 1) (X 0-A 0) (X 1-A 1)

private theorem coordinate_cr
    (A B P Q R : EuclideanSpace ℝ (Fin 2)) :
    cr (cx A B P-cx A B Q) (cy A B P-cy A B Q)
        (cx A B R-cx A B Q) (cy A B R-cy A B Q) =
      dt (B 0-A 0) (B 1-A 1) (B 0-A 0) (B 1-A 1) *
        cr (P 0-Q 0) (P 1-Q 1) (R 0-Q 0) (R 1-Q 1) := by
  dsimp [cx, cy, dt, cr]
  ring

private theorem coordinate_dt
    (A B P Q R : EuclideanSpace ℝ (Fin 2)) :
    dt (cx A B P-cx A B Q) (cy A B P-cy A B Q)
        (cx A B R-cx A B Q) (cy A B R-cy A B Q) =
      dt (B 0-A 0) (B 1-A 1) (B 0-A 0) (B 1-A 1) *
        dt (P 0-Q 0) (P 1-Q 1) (R 0-Q 0) (R 1-Q 1) := by
  dsimp [cx, cy, dt, cr]
  ring

private theorem coordinate_sq
    (A B P Q : EuclideanSpace ℝ (Fin 2)) :
    (cx A B P-cx A B Q)^2+(cy A B P-cy A B Q)^2 =
      dt (B 0-A 0) (B 1-A 1) (B 0-A 0) (B 1-A 1) *
        ((P 0-Q 0)^2+(P 1-Q 1)^2) := by
  dsimp [cx, cy, dt, cr]
  ring

private theorem coordinate_tangent
    (A B P Q R S T U : EuclideanSpace ℝ (Fin 2))
    (h :
      cr (P 0-Q 0) (P 1-Q 1) (R 0-Q 0) (R 1-Q 1) *
          dt (S 0-T 0) (S 1-T 1) (U 0-T 0) (U 1-T 1) -
        dt (P 0-Q 0) (P 1-Q 1) (R 0-Q 0) (R 1-Q 1) *
          cr (S 0-T 0) (S 1-T 1) (U 0-T 0) (U 1-T 1) = 0) :
    cr (cx A B P-cx A B Q) (cy A B P-cy A B Q)
        (cx A B R-cx A B Q) (cy A B R-cy A B Q) *
        dt (cx A B S-cx A B T) (cy A B S-cy A B T)
          (cx A B U-cx A B T) (cy A B U-cy A B T) -
      dt (cx A B P-cx A B Q) (cy A B P-cy A B Q)
        (cx A B R-cx A B Q) (cy A B R-cy A B Q) *
        cr (cx A B S-cx A B T) (cy A B S-cy A B T)
          (cx A B U-cx A B T) (cy A B U-cy A B T) = 0 := by
  rw [coordinate_cr, coordinate_dt, coordinate_dt, coordinate_cr]
  linear_combination
    (dt (B 0-A 0) (B 1-A 1) (B 0-A 0) (B 1-A 1))^2 * h

private theorem coordinate_core
    (b u v x y p q : ℝ)
    (hb : b ≠ 0)
    (hv : v ≠ 0)
    (hKB : (x-b)^2+y^2 ≠ 0)
    (hLC : (p-u)^2+(q-v)^2 ≠ 0)
    (h1 :
      cr (x-b) y (-b) 0 * dt (-u) (-v) (p-u) (q-v) -
      dt (x-b) y (-b) 0 * cr (-u) (-v) (p-u) (q-v) = 0)
    (h2 :
      cr (p-b) q (x-b) y * dt (p-u/2) (q-v/2) (u/2) (v/2) -
      dt (p-b) q (x-b) y * cr (p-u/2) (q-v/2) (u/2) (v/2) = 0)
    (h3 :
      cr (p-u) (q-v) (x-u) (y-v) * dt (b/2) 0 (x-b/2) y -
      dt (p-u) (q-v) (x-u) (y-v) * cr (b/2) 0 (x-b/2) y = 0) :
    (x * q * ((b^2-u^2-v^2)/2) +
     y * (p^2+q^2) * (b-u) +
     (x^2+y^2) * p * (-v) -
     (x^2+y^2) * q * (b-u) -
     y * p * ((b^2-u^2-v^2)/2) -
     x * (p^2+q^2) * (-v)) = 0 := by
  dsimp [dt, cr] at *
  let A1 : ℝ := b*(v+y)
  let B1 : ℝ := 2*(x^2+y^2)-b*(x+u)
  let A2 : ℝ := -b*v-u*y+v*x
  let B2 : ℝ := b*u-u*x-v*y
  let D : ℝ := A1*B2-A2*B1
  have bg8 : b * (A1*(p-u)+B1*(q-v))=0 := by
    dsimp [A1, B1]
    linear_combination -2*h1-4*h3
  have g8 : A1*(p-u)+B1*(q-v)=0 := by
    exact (mul_eq_zero.mp bg8).resolve_left hb
  have bg9 : b * (A2*(p-u)+B2*(q-v))=0 := by
    dsimp [A2, B2]
    linear_combination h1
  have g9 : A2*(p-u)+B2*(q-v)=0 := by
    exact (mul_eq_zero.mp bg9).resolve_left hb
  have hDp : D*(p-u)=0 := by
    dsimp [D]
    linear_combination B2*g8-B1*g9
  have hDq : D*(q-v)=0 := by
    dsimp [D]
    linear_combination A1*g9-A2*g8
  have hD : D=0 := by
    by_contra hnD
    have hp : p-u=0 := (mul_eq_zero.mp hDp).resolve_left hnD
    have hq : q-v=0 := (mul_eq_zero.mp hDq).resolve_left hnD
    apply hLC
    rw [hp, hq]
    norm_num
  have hm :
      v * ((x-b)^2+y^2) *
      (x * q * ((b^2-u^2-v^2)/2) +
       y * (p^2+q^2) * (b-u) +
       (x^2+y^2) * p * (-v) -
       (x^2+y^2) * q * (b-u) -
       y * p * ((b^2-u^2-v^2)/2) -
       x * (p^2+q^2) * (-v)) = 0 := by
    dsimp [D, A1, B1, A2, B2] at hD
    dsimp [A1, B1, A2, B2] at g8 g9
    linear_combination
      (-(b^3*v - 2*b^2*v*x + 2*b*p*u*y + 2*b*q*v*y +
          b*v*x^2 + b*v*y^2 - 2*p*u*x*y - 2*p*v*y^2 +
          2*q*u*y^2 - 2*q*v*x*y)/2)*g8 +
      (-(b^3*v + 2*b^3*y - 3*b^2*u*y + b^2*v*x -
          4*b^2*x*y + 2*b*p*u*y + 2*b*p*x*y + 2*b*q*v*y +
          2*b*q*y^2 + b*u^2*y + 2*b*u*x*y + b*v^2*y -
          4*b*v*x^2 - 2*b*v*y^2 + 4*b*x^2*y + 4*b*y^3 -
          4*p*x^2*y - 4*p*y^3 - 2*u*x^2*y - 2*u*y^3 +
          2*v*x^3 + 2*v*x*y^2)/2)*g9 +
      ((-b^2*v + b*u*y + b*v*x - u^2*y - v^2*y)/2)*hD +
      (2*(b^2*y - b*u*y + b*v*x - 2*b*x*y -
          v*x^2 - v*y^2 + 2*x^2*y + 2*y^3))*h2
  have hn : v * ((x-b)^2+y^2) ≠ 0 := mul_ne_zero hv hKB
  exact (mul_eq_zero.mp hm).resolve_left hn

theorem Midpoints_Interior_AKL :
    ∀ A : EuclideanSpace ℝ (Fin 2),
    ∀ B : EuclideanSpace ℝ (Fin 2),
    ∀ C : EuclideanSpace ℝ (Fin 2),
    AffineIndependent (P := EuclideanSpace ℝ (Fin 2)) ℝ ![A, B, C] →
    let M := midpoint (P := EuclideanSpace ℝ (Fin 2)) ℝ A B;
    let N := midpoint (P := EuclideanSpace ℝ (Fin 2)) ℝ A C;
    ∀ K : EuclideanSpace ℝ (Fin 2),
    ∀ L : EuclideanSpace ℝ (Fin 2),
    AffineIndependent (P := EuclideanSpace ℝ (Fin 2)) ℝ ![B, M, C] →
    AffineIndependent (P := EuclideanSpace ℝ (Fin 2)) ℝ ![B, N, C] →
    K ∈ convexHull ℝ ({B, M, C} : Set (EuclideanSpace ℝ (Fin 2))) →
    L ∈ convexHull ℝ ({B, N, C} : Set (EuclideanSpace ℝ (Fin 2))) →
    AffineIndependent (P := EuclideanSpace ℝ (Fin 2)) ℝ ![A, B, L] →
    AffineIndependent (P := EuclideanSpace ℝ (Fin 2)) ℝ ![A, K, C] →
    K ∈ convexHull ℝ ({A, B, L} : Set (EuclideanSpace ℝ (Fin 2))) →
    L ∈ convexHull ℝ ({A, K, C} : Set (EuclideanSpace ℝ (Fin 2))) →
    (K ≠ B ∧ B ≠ A) →
    (A ≠ C ∧ C ≠ L) →
    (L ≠ B ∧ B ≠ K) →
    (L ≠ N ∧ N ≠ C) →
    (L ≠ C ∧ C ≠ K) →
    (B ≠ M ∧ M ≠ K) →
    (EuclideanGeometry.angle (P := EuclideanSpace ℝ (Fin 2)) K B A) =
      (EuclideanGeometry.angle (P := EuclideanSpace ℝ (Fin 2)) A C L) →
    (EuclideanGeometry.angle (P := EuclideanSpace ℝ (Fin 2)) L B K) =
      (EuclideanGeometry.angle (P := EuclideanSpace ℝ (Fin 2)) L N C) →
    (EuclideanGeometry.angle (P := EuclideanSpace ℝ (Fin 2)) L C K) =
      (EuclideanGeometry.angle (P := EuclideanSpace ℝ (Fin 2)) B M K) →
    AffineIndependent (P := EuclideanSpace ℝ (Fin 2)) ℝ ![A, K, L] →
    ∀ O : EuclideanSpace ℝ (Fin 2),
    (Dist.dist (α := EuclideanSpace ℝ (Fin 2)) O A) =
      (Dist.dist (α := EuclideanSpace ℝ (Fin 2)) O K) →
    (Dist.dist (α := EuclideanSpace ℝ (Fin 2)) O A) =
      (Dist.dist (α := EuclideanSpace ℝ (Fin 2)) O L) →
    ((Dist.dist (α := EuclideanSpace ℝ (Fin 2)) O M) =
      (Dist.dist (α := EuclideanSpace ℝ (Fin 2)) O N)) := by
  intro A B C hABC
  dsimp only
  intro K L _ _ hK₁ hL₁ hABL hAKC hK₂ hL₂
    hKB_A hAC_L hLB_K hLN_C hLC_K hBM_K ha1 ha2 ha3 hAKL O hOA_K hOA_L
  let M := midpoint ℝ A B
  let N := midpoint ℝ A C
  obtain ⟨sg1, sg2, sg3⟩ :=
    three_angle_cross_signs A B C K L hABC hABL hAKC hK₁ hL₁ hK₂ hL₂
  have t1 := angle_tangent_of_same_sign K B A A C L
    hKB_A.1 hKB_A.2.symm hAC_L.1 hLC_K.1 ha1 sg1
  have t2 := angle_tangent_of_same_sign L B K L N C
    hLB_K.1 hLB_K.2.symm hLN_C.1 hLN_C.2.symm ha2 sg2
  have t3 := angle_tangent_of_same_sign L C K B M K
    hLC_K.1 hLC_K.2.symm hBM_K.1 hBM_K.2.symm (by simpa [M] using ha3) (by simpa [M] using sg3)
  have ct1 := coordinate_tangent A B K B A A C L t1
  have ct2 := coordinate_tangent A B L B K L N C t2
  have ct3 := coordinate_tangent A B L C K B M K t3
  let b := cx A B B
  let u := cx A B C
  let v := cy A B C
  let x := cx A B K
  let y := cy A B K
  let p := cx A B L
  let q := cy A B L
  have hb : b ≠ 0 := by
    have hBA := hKB_A.2
    rw [ne_eq, ToPoly.eq'] at hBA
    intro hz
    apply hBA
    dsimp [b, cx, dt] at hz
    linear_combination hz
  have hv : v ≠ 0 := by
    simpa [v, cy, ori] using ori_ne_zero_of_affineIndependent A B C hABC
  have hKB' : (x-b)^2+y^2 ≠ 0 := by
    have hne := hKB_A.1
    rw [ne_eq, ToPoly.eq'] at hne
    intro hz
    apply hne
    have hid :
        (x-b)^2+y^2 =
          b*((K 0-B 0)^2+(K 1-B 1)^2) := by
      dsimp [x, y, b, cx, cy, dt, cr]
      ring
    rw [hz] at hid
    exact (mul_eq_zero.mp hid.symm).resolve_left hb
  have hLC' : (p-u)^2+(q-v)^2 ≠ 0 := by
    have hne := hLC_K.1
    rw [ne_eq, ToPoly.eq'] at hne
    intro hz
    apply hne
    have hid :
        (p-u)^2+(q-v)^2 =
          b*((L 0-C 0)^2+(L 1-C 1)^2) := by
      dsimp [p, q, u, v, b, cx, cy, dt, cr]
      ring
    rw [hz] at hid
    exact (mul_eq_zero.mp hid.symm).resolve_left hb
  have ch1 :
      cr (x-b) y (-b) 0 * dt (-u) (-v) (p-u) (q-v) -
      dt (x-b) y (-b) 0 * cr (-u) (-v) (p-u) (q-v) = 0 := by
    dsimp [x, y, p, q, u, v, b, cx, cy, dt, cr] at ct1 ⊢
    linear_combination ct1
  have ch2 :
      cr (p-b) q (x-b) y * dt (p-u/2) (q-v/2) (u/2) (v/2) -
      dt (p-b) q (x-b) y * cr (p-u/2) (q-v/2) (u/2) (v/2) = 0 := by
    dsimp [x, y, p, q, u, v, b, N, cx, cy, dt, cr] at ct2 ⊢
    to_poly at ct2
    linear_combination ct2
  have ch3 :
      cr (p-u) (q-v) (x-u) (y-v) * dt (b/2) 0 (x-b/2) y -
      dt (p-u) (q-v) (x-u) (y-v) * cr (b/2) 0 (x-b/2) y = 0 := by
    dsimp [x, y, p, q, u, v, b, M, cx, cy, dt, cr] at ct3 ⊢
    to_poly at ct3
    linear_combination ct3
  have hF := coordinate_core b u v x y p q hb hv hKB' hLC' ch1 ch2 ch3
  let ox := cx A B O
  let oy := cy A B O
  have eK : 2*ox*x+2*oy*y-(x^2+y^2)=0 := by
    have hh := hOA_K
    to_poly at hh
    have hcO := coordinate_sq A B O A
    have hcK := coordinate_sq A B O K
    have heq :
        (cx A B O-cx A B A)^2+(cy A B O-cy A B A)^2 =
          (cx A B O-cx A B K)^2+(cy A B O-cy A B K)^2 := by
      rw [hcO, hcK, hh]
    dsimp [ox, oy, x, y, cx, cy, dt, cr] at heq ⊢
    linear_combination heq
  have eL : 2*ox*p+2*oy*q-(p^2+q^2)=0 := by
    have hh := hOA_L
    to_poly at hh
    have hcO := coordinate_sq A B O A
    have hcL := coordinate_sq A B O L
    have heq :
        (cx A B O-cx A B A)^2+(cy A B O-cy A B A)^2 =
          (cx A B O-cx A B L)^2+(cy A B O-cy A B L)^2 := by
      rw [hcO, hcL, hh]
    dsimp [ox, oy, p, q, cx, cy, dt, cr] at heq ⊢
    linear_combination heq
  have hdet : x*q-y*p ≠ 0 := by
    have hori := ori_ne_zero_of_affineIndependent A K L hAKL
    have hc := coordinate_cr A B K A L
    have hxy : x*q-y*p = b*ori A K L := by
      dsimp [x, y, p, q, b, cx, cy, dt, cr, ori] at hc ⊢
      linear_combination hc
    intro hz
    apply hori
    have hp : b * ori A K L = 0 := by
      rw [← hxy, hz]
    exact (mul_eq_zero.mp hp).resolve_left hb
  have eW :
      2*ox*(b-u)+2*oy*(-v)-(b^2-u^2-v^2)/2=0 := by
    apply (mul_eq_zero.mp ?_).resolve_left hdet
    linear_combination
      (x*(-v)-y*(b-u))*eL -
      (p*(-v)-q*(b-u))*eK - hF
  have hscalar :
      (ox-b/2)^2+oy^2 = (ox-u/2)^2+(oy-v/2)^2 := by
    linear_combination (-1/2)*eW
  have cxM : cx A B M = b/2 := by
    dsimp [M, b, cx, dt]
    to_poly
    ring
  have cyM : cy A B M = 0 := by
    dsimp [M, cy, cr]
    to_poly
    ring
  have cxN : cx A B N = u/2 := by
    dsimp [N, u, cx, dt]
    to_poly
    ring
  have cyN : cy A B N = v/2 := by
    dsimp [N, v, cy, cr]
    to_poly
    ring
  have hcoord :
      (cx A B O-cx A B M)^2+(cy A B O-cy A B M)^2 =
        (cx A B O-cx A B N)^2+(cy A B O-cy A B N)^2 := by
    rw [cxM, cyM, cxN, cyN]
    simpa [ox, oy] using hscalar
  have hcM := coordinate_sq A B O M
  have hcN := coordinate_sq A B O N
  have hprod :
      b * (((O 0-M 0)^2+(O 1-M 1)^2) -
        ((O 0-N 0)^2+(O 1-N 1)^2)) = 0 := by
    change
      dt (B 0-A 0) (B 1-A 1) (B 0-A 0) (B 1-A 1) *
        (((O 0-M 0)^2+(O 1-M 1)^2) -
          ((O 0-N 0)^2+(O 1-N 1)^2)) = 0
    rw [mul_sub, ← hcM, ← hcN]
    exact sub_eq_zero.mpr hcoord
  have hz := (mul_eq_zero.mp hprod).resolve_left hb
  to_poly
  dsimp [M, N] at hz
  to_poly at hz
  nlinarith
\end{leancode}

\end{document}